\documentclass[final,12pt]{colt2024_copy} 

\title[DDRM for Laplace Operator]{Denoising Diffusion Restoration Tackles Forward and Inverse Problems for the Laplace Operator}
\usepackage{times}
\usepackage{float}
\usepackage{subcaption}
\newcommand{\R}{\mathbb{R}}

\coltauthor{\Name{Amartya Mukherjee} \Email{a29mukhe@uwaterloo.ca}\\
 \addr Department of Applied Mathematics, University of Waterloo\\
 \Name{Melissa M. Stadt} \\
 \addr Department of Applied Mathematics, University of Waterloo\\
 \Name{Lena Podina} \\
 \addr Department of Computer Science, University of Waterloo\\
 \Name{Mohammad Kohandel} \\
 \addr Department of Applied Mathematics, University of Waterloo\\
 \Name{Jun Liu} \\
 \addr Department of Applied Mathematics, University of Waterloo}

\begin{document}

\maketitle

\begin{abstract}%
Diffusion models have emerged as a promising class of generative models that map noisy inputs to realistic images. More recently, they have been employed to generate solutions to partial differential equations (PDEs). However, they still struggle with inverse problems in the Laplacian operator, for instance, the Poisson equation, because the eigenvalues that are large in magnitude amplify the measurement noise. This paper presents a novel approach for the inverse and forward solution of PDEs through the use of denoising diffusion restoration models (DDRM). DDRMs were used in linear inverse problems to restore original clean signals by exploiting the singular value decomposition (SVD) of the linear operator. Equivalently, we present an approach to restore the solution and the parameters in the Poisson equation by exploiting the eigenvalues and the eigenfunctions of the Laplacian operator. Our results show that using denoising diffusion restoration significantly improves the estimation of the solution and parameters. Our research, as a result, pioneers the integration of diffusion models with the principles of underlying physics to solve PDEs.
\end{abstract}


\section{Introduction}
Denoising diffusion models are among the current leading methods for generative modeling \citep{ho2020denoising, song2020denoising}. They have shown great success in applications such as the generation of images, speech, and video, as well as image super-resolution \citep{song2020denoising, yang_diffusion_2023}. Other applications include physics-guided human motion (e.g., PhysDiff \citep{yuan_physdiff_2023}), customized ODE solvers that are more efficient than Runge-Kutta methods \citep{lu_dpm-solver_2022}, molecule generation \citep{pmlr-v162-hoogeboom22a}, and more \citep{yang_diffusion_2023}. Furthermore, they are stable to train and are relatively easy to scale \citep{yang_diffusion_2023}. 

The Laplace operator is a differential operator of second order, $\Delta=\nabla\cdot\nabla$ \citep{gilbarg1983elliptic}, which appears in many partial differential equations (PDEs) such as the Poisson equation, heat equation, and wave equation. It is a compact self-adjoint operator and thus, has an orthonormal set of eigenfunctions and real eigenvalues \citep{chavel1984eigenvalues}. This paper concerns two problems involving the Laplace operator defined in the domain $\Omega=[0,1]^2$; (i) the forward problem, where we are given a function $f\in C(\Omega)$ and we intend to compute $u$ satisfying $\Delta u=f$ and $u=0$ in the boundary, $\partial\Omega$, and (ii) the inverse problem, where we are given a function $u\in C^2(\Omega)$ with $u=0$ in $\partial\Omega$, and we intend to compute $\Delta u$.

Since many PDEs do not have an analytical solution, numerical methods are necessary for obtaining solutions to these systems \citep{thomas_numerical_2013, quarteroni_numerical_2006}. However, numerical methods lead to known numerical errors and are often computationally expensive, especially for complex PDEs. Additionally, while there are many PDE solvers, they are often restricted to the specific type of PDE they are designed for. When working to understand the physics of a system, these numerical errors add noise which makes the physics difficult to solve. In recent years, deep learning techniques have been introduced to solve PDEs. Such examples include physics-informed neural networks (PINNs) \citep{raissi2019physics}, deep operator networks (DeepONets) \citep{lu2019deeponet}, and Fourier neural operators (FNOs) \citep{li2020fourier}. These techniques have been used to improve computational efficiency, conduct reduced-order modeling, and develop generalized PDE solvers. Although these techniques are effective in solving PDEs, their precision and generalizability, like many machine and deep learning methods, are limited by the scarcity of high-quality training data.

In more recent works, \citealt{apte2023diffusion} seek to address the problem of data scarcity for machine learning methods of PDE modeling by developing a method of data generation using a diffusion model. By training on data from the steady 2D Poisson equation on a fixed square domain, they made diffusion models generate paired data samples that adhered to physics laws, despite not including physics in the model directly. They trained a model to generate pairs of $u$ and $f$ satisfying $\Delta u=f$, thus addressing the challenge of capturing the joint distribution of $u$ and $f$. Their analysis, however, only concerns pairs of $u$ and $f$ that the model generates without imposing any constraints. Frequently, we are tasked with either being given the solution or the parameters of a PDE and are required to estimate the counterpart.
We intend to address this challenge by introducing conditional generation into this model. Our goal is to solve the forward problem by sampling $u$ from the model by fixing the parameters $f$ as a constraint and to solve the inverse problem by sampling $f$ from the model by fixing the solution $u$ as a constraint.
Furthermore, contrary to past works, we exploit the physics of the Laplace operator to derive eigenvalue and eigenfunction pairs that we project $u$ and $f$ onto to improve the sampling.

We first start by replicating diffusion model-based data generation for the 2D Poisson equation with homogeneous Dirichlet boundary conditions as in \citealt{apte2023diffusion}. We trained a denoising diffusion implicit model (DDIM) \citep{song2020denoising} on a sample of 38,250 data points. After training the diffusion model, we generate numerical solutions $u(x,y)$ conditioned on the parameter $f(x,y)$, which we will refer to as the forward process. Our numerical results are posted in Appendix \ref{app:inverse_u}, and show that the DDIM model is a great, albeit noisy numerical solver to the Poisson equation. We attempt to generate approximations to the parameter $f(x,y)$ conditioned on $u(x,y)$, which we will refer to as the inverse process. Our numerical results are posted in Appendix \ref{app:forward_f} and demonstrate that the DDIM model is a poor numerical solver for the inverse problem. Therefore, we require a better method to solve the inverse problem.

To solve this problem, we employ denoising diffusion restoration models (DDRMs) \citep{kawar2022denoising,chung2023diffusion,murata2023gibbsddrm}. DDRMs are used to restore clean data in linear inverse problems using a pre-trained diffusion model without requiring any fine-tuning. The authors achieve this by using the singular value decomposition (SVD) of the linear operator to transform the original signal and observed signal to a shared spectral space. DDRMs showed state-of-the-art performance in restoring realistic images in super-resolution and deblurring tasks by assuming that the original clean image is returned by a generative model.

\begin{figure}[H]
    \centering
     \subfigure[Forward Problem - Restoring $u$ conditioned on $f$ with DDRM]{\includegraphics[width=0.9\linewidth,trim={3cm 0 3cm 1cm},clip]{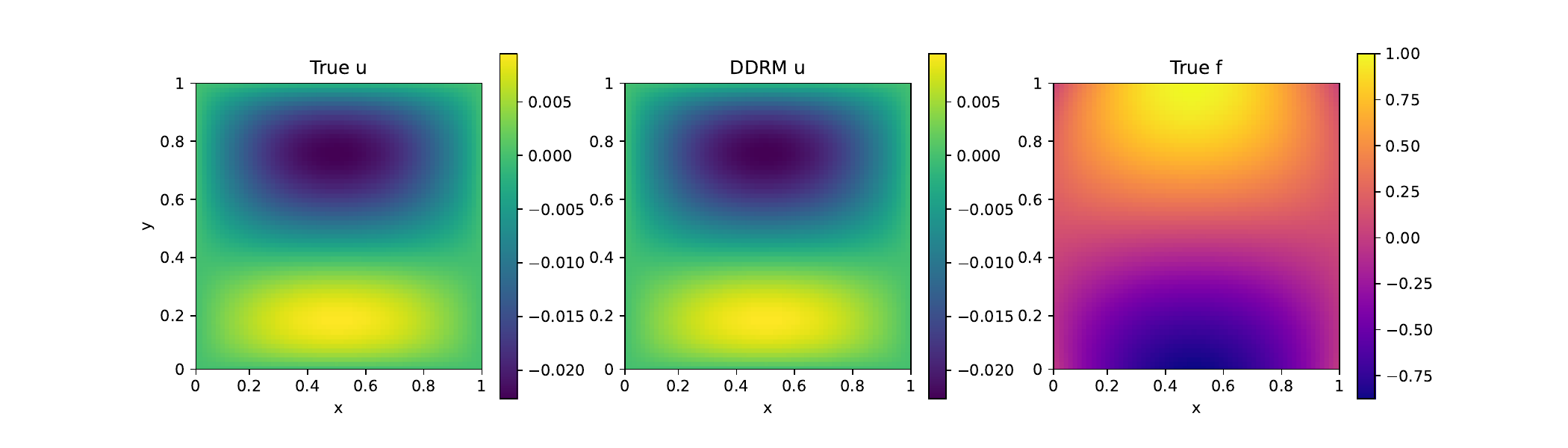}}
     \subfigure[Inverse Problem - Restoring $f$ conditioned on $u$ with DDRM]{\includegraphics[width=0.9\linewidth,trim={3cm 0 3cm 1cm},clip]{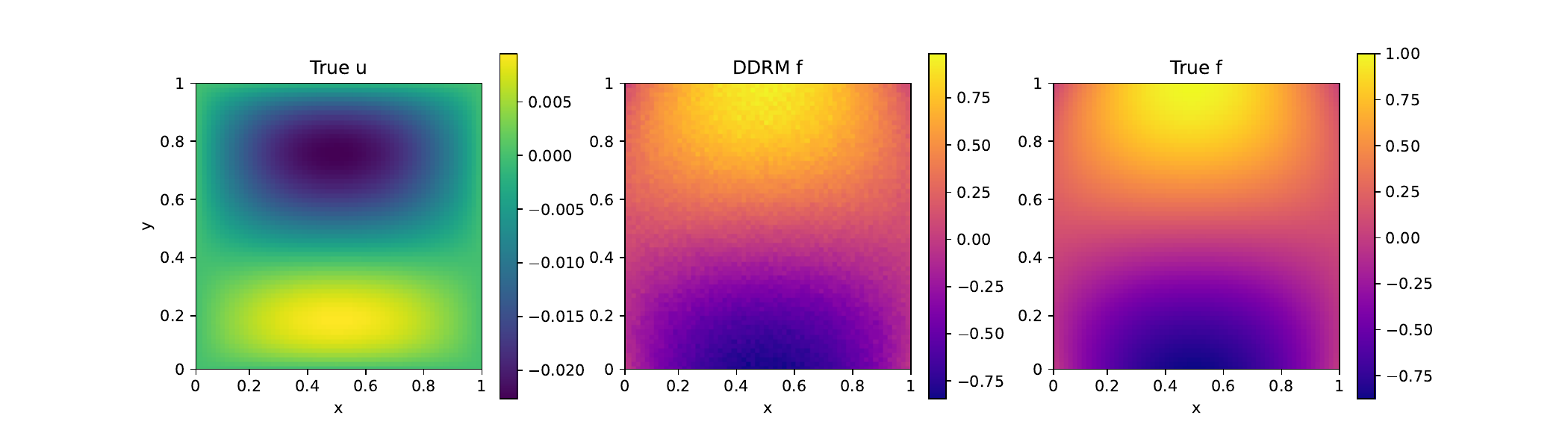}}
        \caption{Plots of restored solutions of the forward problem $u(x,y)$ and inverse problem $f(x,y)$ (middle), provided analytical solutions $u(x,y)$ (left) and $f(x,y)$ (right).}\label{fig:intro_fig}
\end{figure}

We use DDRMs to solve the forward and inverse problem based on \cite{kawar2022denoising}. Similar to how DDRMs solve linear inverse problems by exploiting the singular value decomposition of the linear operator, we solve forward and inverse problems in the Poisson equation by exploiting the eigenspace of the Laplace operator constrained to homogeneous Dirichlet boundary conditions. Our method shows a significant improvement in the restoration of the parameters, achieving a mean absolute error (MAE) of $3.215\times 10^{-2}$. Furthermore, we achieve an average MAE of $1.175\times 10^{-6}$ in our improved forward process, which is just slightly greater than the MAE of $6.672\times 10^{-7}$ upon using the finite difference method, thus showing a significant improvement by using DDRM. Our results are briefly shown in Figure \ref{fig:intro_fig}. Our work outperforms other data-driven benchmarks such as PINNs and DeepONets and is the first to do so by including the physics in diffusion models.

\section{Problem statement and conditional distributions}


We will start by describing the inverse and forward problems associated with the Laplacian operator by considering the Poisson equation defined on two spatial dimensions
\begin{equation}\label{eq:Poisson2D}
    \Delta u=f, 
\end{equation}
where the domain is $\Omega=[0,1]^2$ with homogenous Dirichlet boundary conditions $u=0$ on $\partial\Omega$. The inverse problem concerns computing $f$ conditioned on $u$, and the forward problem concerns computing $u$ conditioned on $f$ -- in other words, solving the PDE.

\subsection*{Forward problem}

Consider the following forward problem defined on a domain $\Omega=[0,1]^2$: 
\begin{equation}
    f=\Delta u_0+z_f,\label{eq:Blindforward}
\end{equation}
where $z_f\sim \mathcal{N}(0,\sigma_u^2)$ is measurement noise with known covariance $\sigma_u^2$,  $\Delta u_0$ is the Laplacian of a function $u_0\in C^2(\Omega)\cup C^1(\partial\Omega)$, and $f_0\in C(\Omega)$ is a forcing function. We also have the boundary conditions $u_0=0$ on $\partial\Omega$. As shown in Appendix \ref{app:inverse_u}, a pre-trained diffusion model is effective in retrieving the solution $u_0(x,y)$ while keeping the parameter channel $f(x,y)$ fixed, which we will refer to as a dry forward process. However, there is a significant amount of noise in these estimated solutions, thus increasing the MAE with respect to the true solution.

\subsection*{Inverse problem}

Consider the following inverse problem defined on a domain $\Omega=[0,1]^2$: 
\begin{align}
    f_0&=\Delta u_0,\nonumber\\
    u&=u_0+z_u,\label{eq:Blindinverse}
\end{align}
where $\Delta u_0$ is the Laplacian of a function $u_0\in C^2(\Omega)\cup C^1(\partial\Omega)$, and $f_0\in C(\Omega)$ is a forcing function. We have the boundary conditions $u_0=0$ on $\partial\Omega$,  $z_u(x,y)$ is a Brownian bridge satisfying the same boundary conditions as $u_0$, and $u$ is the observed signal with measurement noise from which we need to estimate the parameter $f_0$.
As shown in Appendix \ref{app:forward_f}, a pre-trained diffusion model fails to retrieve the parameter $f_0(x,y)$ while keeping the solution channel $u(x,y)$ fixed, which we will refer to as a dry inverse process.

In this section, we will introduce some prior work that assists us in deriving an improved method to restore $f(x,y)$ and $u(x,y)$.

\subsection{Denoising diffusion restoration models}

DDRM is a method that uses a pre-trained diffusion model $p_\theta$ as a prior for data \citep{kawar2022denoising}. It is used to restore clean images in non-blind linear inverse problems of the form $y=Hx_0+z$, where $x_0$ is the original image, $z$ is measurement noise with known covariance, and $H$ is a linear operator. DDRM is defined as a Markov chain $x_T\to x_{T-1}\to\dots\to x_1\to x_0$ conditioned on $y$:
\begin{equation}
    p(x_{0:T}\mid y)=p_{\theta}^{(T)}(x_T\mid y)\prod_{t=0}^{T-1}p_\theta^{(t)}(x_t\mid x_{t+1},y).
\end{equation}
DDRM uses the singular value decomposition of $H$ to project $x_T$ and $y$ into a shared spectral space.
It has shown improved performance in restoring clean images in multiple tasks such as image deblurring, inpainting removal, image coloration, and super-resolution \citep{kawar2022denoising}.
This work has been followed by the works of \citep{chung2023diffusion,murata2023gibbsddrm} that extend DDRMs to blind forward problems where the operator $H$ is unknown.

\subsection{Eigenvalues and eigenfunctions of the Laplace operator}

Although problems like equation \eqref{eq:Blindinverse} do not have a notion of singular value decomposition, its eigenvalue decomposition has been explored in past works in PDE literature and offers us a method to project $u$ and $f$ into a shared spectral space. That is explained by the following proposition.
\begin{proposition}\label{thm:eigenpow}
    The eigenfunction and eigenvalue pairs of the Laplacian operator $\Delta$ in a domain $\Omega=[0,1]^2$ subject to the boundary conditions $u=0$ on $\partial\Omega$ are of the form
    \begin{align}
        &u_{n,m}(x,y)=\sin(n\pi x)\sin(m\pi y),\\&\lambda_{n,m}=-(n\pi)^2-(m\pi)^2.
    \end{align}
\end{proposition}
For completeness, a proof of this known proposition is provided in Appendix \ref{app:pf_eigenpow}. This proposition will be useful in our derivation of the modified DDRM algorithm. We are projecting the functions $u$ and $f$ into the eigenfunctions of the Laplacian operator, thus allowing us to solve them on a shared spectral space. The proposition also explains the numerical results shown in Appendix \ref{app:forward_f} and \ref{app:inverse_u}. Due to the large magnitude of the eigenvalues, the Laplacian operator minimizes the measurement noise in the forward process, but it amplifies the noise measurement noise in the inverse process, thus making $f_0$ harder to compute.

\subsection{Introducing conditional distributions}

In the forward and inverse problem with the Laplacian operator, it is important to note that assuming that the measurement noise is i.i.d along the domain may not be realistic. Recognizing that the distribution of the noise varies along $x$ and $y$ is important to sample solutions that are consistent with our knowledge of the problem, such as the PDE and the boundary conditions. Naturally, it makes sense that our solution has the highest uncertainty along the center of the domain (i.e. points near $(0.5,0.5)$).

\subsubsection{Forward problem}\label{sec:condition_dist_forward}

We define the noise $z_f$ as i.i.d Gaussian since we do not impose any assumptions on $f(x,y)$ other than continuity. This, however, does not apply to $u(x,y)$. We introduce Theorem \ref{thm:greens_distribution} that models its distribution along $(x,y)$.
\begin{theorem}\label{thm:greens_distribution}
If $f=\Delta u+z_f$ where $z_f\sim \mathcal{N}(0,\sigma_f^2)$, then the marginal distribution of $u$ conditioned on $f$ is
\begin{equation}
    u(x,y)\mid f_0\sim \mathcal{N}(u_0,\sigma_f^2K(x,y)),
\end{equation}
where
\begin{equation}\label{eq:K}
    K(x,y)=\iint_{\Omega}(\psi((x',y')-(x,y)))^2dx'dy',
\end{equation}
where $\psi(\cdot)$ is the Green's function in two dimensions
\begin{equation}
    \psi(x,y)=\frac{\ln(\|(x,y)\|)}{2\pi}. 
\end{equation}
\end{theorem}
A proof of the theorem is provided in Appendix \ref{app:pf_greens_distribution}. This theorem will be very important in our DDRM algorithm for sampling $u(x,y)$ conditioned on $f(x,y)$. 
In the development of our algorithm, we are interested in the distribution of the discrete sine transform of $u(x,y)$. Computing this distribution is expensive, so we introduce a theorem that places an upper bound on the variance of the discrete sine transform of $u(x,y)$.
\begin{theorem}\label{thm:dst_u_var}
    Let $f=\Delta u+z_f$ where $z_f\sim \mathcal{N}(0,\sigma_f^2)$. Consider the discrete sine transform (DST) of $u(x,y)$
    \begin{equation}
    \overline{\mathbf{u}}^{(n,m)}=\langle\mathbf{u},\sin(n\pi x)\sin(m\pi y)\rangle.
    \end{equation}
    We can place an upper bound on the variance of $\overline{\mathbf{u}}^{(n,m)}$: 
    \begin{equation}\label{eq:dst_u_var}
        Var[\overline{\mathbf{u}}^{(n,m)}|f_0]\leq\left(\frac{1}{\pi^2(n^2+m^2)}+\ln 2\max(n,m)\right)^2\sigma_f^2. 
    \end{equation}
\end{theorem}
A proof of the theorem is provided in Appendix \ref{app:pf_dst_u_var}. This theorem follows from Theorem \ref{thm:greens_distribution} and helps us define distributions to sample $u(x,y)$ conditioned on $f(x,y)$. Notice that the coefficient of the variance term can be computed analytically, thus significantly reducing the computation cost of our algorithm compared to computing $K(x,y)$ using numerical integration.



\subsubsection{Inverse problem}\label{sec:condition_dist_inverse}

\begin{figure}
    \centering
    \includegraphics[width=\linewidth,trim={3cm 0 3cm 0},clip]{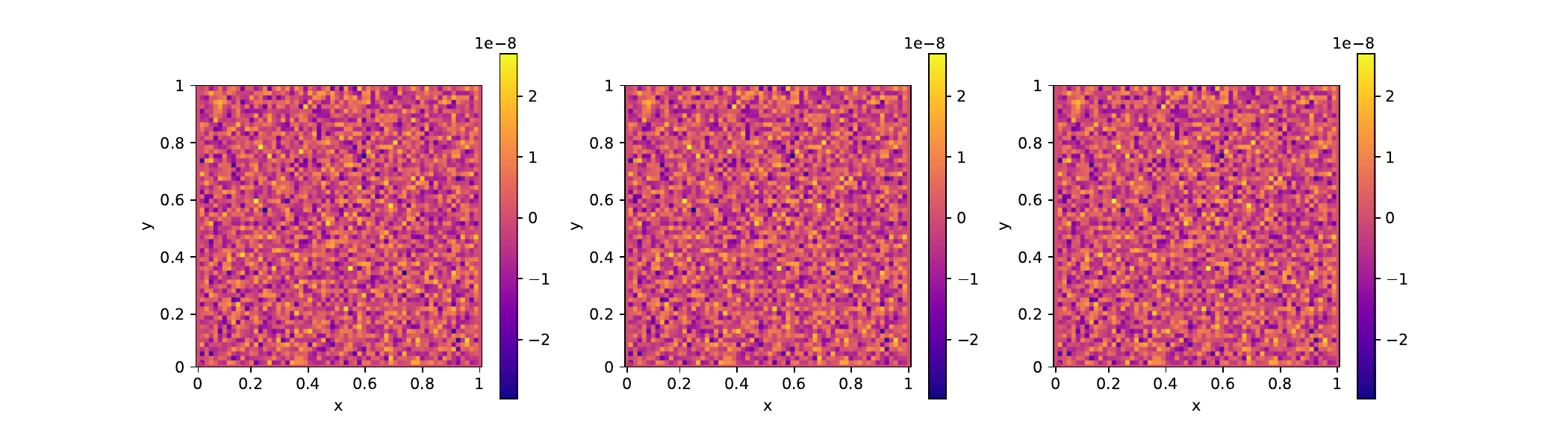}
    \caption{Numerical simulations of Brownian bridge with $\sigma_{n,m}=1\times 10^{-6}$ for all $n,m$}
    \label{fig:brownian_bridge}
\end{figure}

We define the noise $z_u$ as a Brownian bridge satisfying the homogenous boundary conditions $z_u=0$ on $\partial\Omega$. A common approach is to express $z_u$ as a double sum of sinusoidal functions (satisfying the boundary conditions) 
\begin{equation}\label{eq:w_mn}
    z_u=\sum_{n=0}^{N}\sum_{m=0}^{N}w_{n,m}\sin(n\pi x)\sin(m\pi y),
\end{equation}
where $w_{n,m}\sim \mathcal{N}(0,\sigma_{n,m}^2)$ are random coefficients with known variance $\sigma_{n,m}^2$. Putting this together, $z_u$ has the following distribution
\begin{equation}
    z_u(x,y)\sim N\left(0,\sum_{n=0}^{N}\sum_{m=0}^{N}\sigma_{n,m}^2\sin(n\pi x)^2\sin(m\pi y)^2\right).
\end{equation}
Three numerical simulations of this Brownian bridge have been plotted in Figure \ref{fig:brownian_bridge} with $\sigma_{n,m}=10^{-6}$ for all $n,m$.

\section{DDRM for solving PDEs}\label{sec:method}

\subsection{Forward process: Sampling $u$ conditioned on $f$}

In this section, we will explain the use of DDRM in sampling $u$ while conditioned on $f$. We will denote the projection of $f(x,y)$ and $u(x,y)$ into a 64$\times$64 grid as $\mathbf{f}$ and $\mathbf{u}$ respectively. We define the DDRM in this example to be a Markov chain $\mathbf{u}_T\to\mathbf{u}_{T-1}\to\dots\to\mathbf{u}_1\to\mathbf{u}_0$ conditioned on $\mathbf{f}$:
\begin{equation}
    p_{\theta}(\mathbf{u}_{0:T}\mid \mathbf{f})=p_{\theta}^{(T)}(\mathbf{u}_T\mid \mathbf{f})\prod_{t=0}^{T-1}p_\theta^{(t)}(\mathbf{u}_t\mid \mathbf{u}_{t+1},\mathbf{f}).
\end{equation}
In the sampling of $\mathbf{u}_T$ and $\mathbf{u}_t$ for $t=0,...,T-1$, our approach is a modified version of \cite{kawar2022denoising}, where we consider the projection of $u$ and $f$ in the eigenfunctions of the Laplacian operator instead of singular vectors of a linear operator. Similarly to their work, we will consider the variational distribution conditioned on $\mathbf{u}$:
\begin{equation}
    q(\mathbf{f}_{1:T}\mid \mathbf{f}_{0},\mathbf{u})=p_{\theta}^{(T)}(\mathbf{f}_T\mid \mathbf{f}_{0},\mathbf{u})\prod_{t=0}^{T-1}p_\theta^{(t)}(\mathbf{f}_t\mid \mathbf{f}_{t+1},\mathbf{f}_{0},\mathbf{u}).
\end{equation}
In this process, we consider the DST of $\mathbf{u}_t$ and $\mathbf{f}$ and perform the diffusion in its spectral space. Define $\overline{\mathbf{u}}_t^{(n,m)}$ and $\overline{\mathbf{f}}^{(n,m)}$ as follows:
\begin{align}\label{eq:forward_dst_u}
    \overline{\mathbf{u}}_t^{(n,m)}&=\langle\mathbf{u}_t,\sin(n\pi x)\sin(m\pi y)\rangle,\\\label{eq:forward_dst_f}
    \overline{\mathbf{f}}^{(n,m)}&=\langle\mathbf{f},\sin(n\pi x)\sin(m\pi y)\rangle/\lambda_{n,m},
\end{align}
where $\lambda_{n,m}$ are the eigenvalues from Proposition \ref{thm:eigenpow}. 
We also define $\overline{\mathbf{K}}^{(n,m)}$ as the variance term of $\overline{\mathbf{u}}_t^{(n,m)}$ introduced in Theorem \ref{thm:dst_u_var}:
\begin{equation}
    \overline{\mathbf{K}}^{(n,m)}=\left(\frac{1}{|\lambda_{n,m}|}+\ln 2\max(n,m)\right)^2.
\end{equation}
$\overline{\mathbf{K}}^{(n,m)}$ has a maximum of $1967.938$ corresponding to $n=1,m=64$.
Since none of the eigenvalues $\lambda_{m,n}$ are zero, we can define the variational distribution for $\mathbf{u}_T$ for each index $n,m$ in $\overline{\mathbf{u}}_T^{(n,m)}$ as:
\begin{equation}\label{eq:variational_forward_1}
    q^{(T)}(\overline{\mathbf{u}}_T^{(n,m)} \mid \mathbf{u}_0,\mathbf{f})=\mathcal{N}(\overline{\mathbf{f}}^{(n,m)},\sigma_T^2-\sigma_{f}^2\overline{\mathbf{K}}^{(n,m)}/\lambda_{n,m}^2),
\end{equation}
where $\sigma_{f}$ are defined as the standard deviation of $w_{n,m}$ in equation \eqref{eq:w_mn}. We assume that $\sigma_T>\sigma_{f}\sqrt{\overline{\mathbf{K}}^{(n,m)}}/\lambda_{n,m}$ for all $n,m$. We can also define the variational distribution for $\mathbf{u}_t$ for each index $n,m$ in $\overline{\mathbf{u}}_t^{(n,m)}$ as:
\begin{align}\label{eq:variational_forward_t}
    & q^{(t)}\left(\overline{\mathbf{u}}_t^{(n,m)} \mid \mathbf{u}_0,\mathbf{u}_{t+1}, \mathbf{f}\right)= \nonumber\\
    & \begin{cases}
    \mathcal{N}\Big(\overline{\mathbf{u}}_{0}^{(n,m)}+\sqrt{1-\eta^2} \sigma_t \frac{\overline{\mathbf{f}}^{(n,m)}-\overline{\mathbf{u}}_{0}^{(n,m)}}{\sigma_{f}\sqrt{\overline{\mathbf{K}}^{(n,m)}} / \lambda_{n,m}},\eta^2 \sigma_t^2\Big), & \sigma_t<\frac{\sigma_{f}\sqrt{\overline{\mathbf{K}}^{(n,m)}}}{\lambda_{n,m}}, \\
    \mathcal{N}\Big(\left(1-\eta_b\right) \overline{\mathbf{u}}_{0}^{(n,m)}+\eta_b \overline{\mathbf{f}}^{(n,m)},\sigma_t^2-\frac{\sigma_{f}^2\overline{\mathbf{K}}^{(n,m)}}{\lambda_{n,m}^2} \eta_b^2\Big), &\sigma_t \geq \frac{\sigma_{f}\sqrt{\overline{\mathbf{K}}^{(n,m)}}}{\lambda_{n,m}},\end{cases}
\end{align}
where $\eta,\eta_b\in[0,1]$ are hyperparameters controlling the variance of the distributions. We introduce a proposition that verifies the convergence of this variational distribution.
\begin{proposition}\label{prop:variational_forward}  (Modified version of Proposition 3.1 from \cite{kawar2022denoising})
    The conditional distributions defined in equations \eqref{eq:variational_forward_1} and \eqref{eq:variational_forward_t} satisfy the following Gaussian marginal property:
    \begin{equation}
        q^{(t)}\left(\overline{\mathbf{u}}_t^{(n,m)} \mid {\mathbf{u}}_0\right)=\mathcal{N}(\overline{\mathbf{u}}_0^{(n,m)},\sigma_{f}^2\overline{\mathbf{K}}^{(n,m)}). 
    \end{equation}
\end{proposition}
The proof of this proposition is in Appendix \ref{app:pf_variational_forward}. Based on the formulation of the conditional distribution in Section \ref{sec:condition_dist_forward}, this proposition shows that the transitions $\mathbf{u}_t$ converge in distribution to the distribution of $\mathbf{u}_0$.

\subsection*{Sampling of $\mathbf{u}_T$}

The sampling of $\mathbf{u}_T$ is performed by sampling from the distribution $p(\mathbf{u}_T\mid \mathbf{f})$. Sampling from this conditional distribution is intractable, so we use our modified DDRM to approximate the distribution.
Since none of the eigenvalues $\lambda_{m,n}$ are zero, our DDRM method for sampling $\mathbf{u}_T$ is as follows:
\begin{equation}
    p_{\theta}^{(T)}(\overline{\mathbf{u}}_T^{(n,m)} \mid \mathbf{f})=\mathcal{N}(\overline{\mathbf{f}}^{(n,m)},\sigma_T^2-\sigma_{f}^2\overline{\mathbf{K}}^{(n,m)}/\lambda_{n,m}^2). 
\end{equation}
For this sampling to be tractable, we choose an appropriate value for $\sigma_T$ such that $\sigma_T>\sigma_{f}/(\pi\sqrt{2})$. This distribution is identical to the variational distribution from equation \eqref{eq:variational_forward_1}.

\subsection*{Sampling of $\mathbf{u}_t$}

The sampling of $\mathbf{u}_t$ is performed by sampling from the distribution $p(\mathbf{u}_t\mid \mathbf{u}_{t+1},...,\mathbf{u}_{T},\mathbf{f})$. Sampling from this conditional distribution is intractable, so we use our modified DDRM to approximate the distribution. We denote the prediction of $\mathbf{u}_0$ at time step $t$ as $\mathbf{u}_{\theta, t}$. Our DDRM method of sampling $\mathbf{u}_t$ is as follows:
\begin{align}
    & p_\theta^{(t)}\left(\overline{\mathbf{u}}_t^{(n,m)} \mid \mathbf{u}_{t+1}, \mathbf{f}\right)= \nonumber\\
    & \begin{cases}
    \mathcal{N}\Big(\overline{\mathbf{u}}_{\theta, t}^{(n,m)}+\sqrt{1-\eta^2} \sigma_t \frac{\overline{\mathbf{f}}^{(n,m)}-\overline{\mathbf{u}}_{\theta, t}^{(n,m)}}{\sigma_{f}\sqrt{\overline{\mathbf{K}}^{(n,m)}} / \lambda_{n,m}},\eta^2 \sigma_t^2\Big), & \sigma_t<\frac{\sigma_{f}\sqrt{\overline{\mathbf{K}}^{(n,m)}}}{\lambda_{n,m}}, \\
    \mathcal{N}\Big(\left(1-\eta_b\right) \overline{\mathbf{u}}_{\theta, t}^{(n,m)}+\eta_b \overline{\mathbf{f}}^{(n,m)},\sigma_t^2-\frac{\sigma_{f}^2\overline{\mathbf{K}}^{(n,m)}}{\lambda_{n,m}^2} \eta_b^2\Big), &\sigma_t \geq \frac{\sigma_{f}\sqrt{\overline{\mathbf{K}}^{(n,m)}}}{\lambda_{n,m}}.\end{cases}
\end{align}
This distribution is obtained by modifying the variational distribution from equation \eqref{eq:variational_forward_t}. Since $\mathbf{u}_0$ is unknown at time step $t$, we need to use our pre-trained diffusion model to approximate it, which we denote as $\mathbf{u}_{\theta, t}$.

\subsection{Inverse process: Sampling $f$ conditioned on $u$}

We consider the DST of $\mathbf{u}$ and $\mathbf{f}_t$ and perform the diffusion in its spectral space. Define $\overline{\mathbf{u}}^{(n,m)}$ and $\overline{\mathbf{f}}_t^{(n,m)}$ as follows:
\begin{align}\label{eq:inverse_dst_u}
    \overline{\mathbf{u}}^{(n,m)}&=\lambda_{n,m}\langle\mathbf{u},\sin(n\pi x)\sin(m\pi y)\rangle,\\\label{eq:inverse_dst_f}
    \overline{\mathbf{f}}_t^{(n,m)}&=\langle\mathbf{f}_t,\sin(n\pi x)\sin(m\pi y)\rangle,
\end{align}
where $\lambda_{n,m}$ are the eigenvalues from Proposition \ref{thm:eigenpow}.
To clarify the distinction from the definitions in \eqref{eq:forward_dst_u},\eqref{eq:forward_dst_f}, note which variable has a subscript of $t$.
Since none of the eigenvalues $\lambda_{m,n}$ are zero, we can defined the variational distribution for $\mathbf{f}_T$ for each index $n,m$ in $\overline{\mathbf{f}}_T^{(n,m)}$ as:
\begin{equation}\label{eq:variational_inverse_1}
    q^{(T)}(\overline{\mathbf{f}}_T^{(n,m)} \mid {\mathbf{f}}_0, \mathbf{u})=\mathcal{N}(\overline{\mathbf{u}}^{(n,m)},\sigma_T^2-\sigma_{n,m}^2\lambda_{n,m}^2),
\end{equation}
where $\sigma_{n,m}$ are defined as the standard deviation of $w_{n,m}$ in equation \eqref{eq:w_mn}. We assume that $\sigma_T>\sigma_{m,n}\lambda_{n,m}$ for all $n,m$. We can also define the variational distribution for $\mathbf{f}_t$ for each index $n,m$ in $\overline{\mathbf{f}}_t^{(n,m)}$ as:
\begin{align}\label{eq:variational_inverse_t}
    & q^{(t)}\left(\overline{\mathbf{f}}_t^{(n,m)} \mid {\mathbf{f}}_0,\mathbf{f}_{t+1}, \mathbf{u}\right)= \nonumber\\
    & \begin{cases}
    \mathcal{N}\Big(\overline{\mathbf{f}}_{0}^{(n,m)}+\sqrt{1-\eta^2} \sigma_t \frac{\overline{\mathbf{u}}^{(n,m)}-\overline{\mathbf{f}}_{0}^{(n,m)}}{\sigma_{n,m} \lambda_{n,m}},\eta^2 \sigma_t^2\Big), & \text{if } \sigma_t<\sigma_{n,m}{\lambda_{n,m}}, \\
    \mathcal{N}\Big(\left(1-\eta_b\right) \overline{\mathbf{f}}_{0}^{(n,m)}+\eta_b \overline{\mathbf{u}}^{(n,m)},\sigma_t^2-{\sigma_{n,m}^2}{\lambda_{m,n}^2} \eta_b^2\Big), & \text{if } \sigma_t \geq \sigma_{n,m}\lambda_{n,m},\end{cases}
\end{align}
where $\eta,\eta_b\in[0,1]$ are hyperparameters controlling the variance of the distributions. We introduce a proposition that verifies the convergence of this variational distribution.
\begin{proposition}\label{prop:variational_inverse} (Modified version of Proposition 3.1 from \cite{kawar2022denoising})
    The conditional distributions defined in equations \eqref{eq:variational_inverse_1} and \eqref{eq:variational_inverse_t} satisfy the following Gaussian marginal property:
    \begin{equation}
        q^{(t)}\left(\overline{\mathbf{f}}_t^{(n,m)} \mid {\mathbf{f}}_0\right)=\mathcal{N}(\overline{\mathbf{f}}_0^{(n,m)},\sigma_{n,m}^2). 
    \end{equation}
\end{proposition}
The proof of this proposition is in Appendix \ref{app:pf_variational_inverse}. Based on the formulation of the conditional distribution in Section \ref{sec:condition_dist_inverse}, this proposition shows that the transitions $\mathbf{f}_t$ converge in distribution to the distribution of $\mathbf{f}_0$.

\subsection*{Sampling of $\mathbf{f}_T$}

The sampling of $\mathbf{f}_T$ is performed by sampling from the distribution $p(\mathbf{f}_T\mid \mathbf{u})$. Sampling from this conditional distribution is intractable, so we use our modified DDRM to approximate the distribution.
Since none of the eigenvalues $\lambda_{m,n}$ are zero, our DDRM method for sampling $\mathbf{f}_T$ is as follows:
\begin{equation}
    p_{\theta}^{(T)}(\overline{\mathbf{f}}_T^{(n,m)} \mid \mathbf{u})=\mathcal{N}(\overline{\mathbf{u}}^{(n,m)},\sigma_T^2-\sigma_{n,m}^2\lambda_{n,m}^2).
\end{equation}
In this process, we assume that $\sigma_T$ is sufficiently large to satisfy $\sigma_T>\sigma_{n,m}((n\pi)^2+(m\pi)^2)$ for all $n,m\in\{1,...,64\}$, so that the variance term is non-negative. This distribution is identical to the variational distribution from \eqref{eq:variational_inverse_1}.

\subsection*{Sampling of $\mathbf{f}_t$}

The sampling of $\mathbf{f}_t$ is performed by sampling from the distribution $p(\mathbf{f}_t\mid \mathbf{f}_{t+1},...,\mathbf{f}_{T},\mathbf{u})$. Sampling from this conditional distribution is intractable, so we use our modified DDRM to approximate the distribution. We denote the prediction of $\mathbf{f}_0$ at time step $t$ as $\mathbf{f}_{\theta, t}$. Our DDRM method of sampling $\mathbf{f}_t$ is as follows:
\begin{align}
    & p_\theta^{(t)}\left(\overline{\mathbf{f}}_t^{(n,m)} \mid \mathbf{f}_{t+1}, \mathbf{u}\right)= \nonumber\\
    & \begin{cases}
    \mathcal{N}\Big(\overline{\mathbf{f}}_{\theta, t}^{(n,m)}+\sqrt{1-\eta^2} \sigma_t \frac{\overline{\mathbf{u}}^{(n,m)}-\overline{\mathbf{f}}_{\theta,t}^{(n,m)}}{\sigma_{n,m} \lambda_{n,m}},\eta^2 \sigma_t^2\Big), & \text{if } \sigma_t<\sigma_{n,m}{\lambda_{n,m}}, \\
    \mathcal{N}\Big(\left(1-\eta_b\right) \overline{\mathbf{f}}_{\theta, t}^{(n,m)}+\eta_b \overline{\mathbf{u}}^{(n,m)},\sigma_t^2-\sigma_{n,m}^2\lambda_{m,n}^2 \eta_b^2\Big), & \text{if } \sigma_t \geq \sigma_{n,m}\lambda_{n,m},\end{cases}
\end{align}
where $0 \leq \eta \leq 1$ and $0 \leq \eta_b \leq 1$ are hyperparameters, and $0=\sigma_0<\sigma_1<\sigma_2<\cdots<\sigma_T$ are noise levels that is the same as that defined with the pre-trained diffusion model. This distribution is obtained by modifying the variational distribution from \eqref{eq:variational_inverse_t}. Since $\mathbf{f}_0$ is unknown at time step $t$, we need to use our pre-trained diffusion model to approximate it, which we denote as $\mathbf{f}_{\theta,t}$.

\section{Implementation}

Before using DDRM to solve the inverse and forward problems, we had to train a diffusion model on a dataset of solutions of the Poisson equation. We replicated the method of PDE solution generation of the 2D Poisson equation with Dirichlet boundary conditions as done in \cite{apte2023diffusion}. In this section, we explain the dataset generation and the training of this model.

\subsection{Dataset generation}\label{sec:data}

To train the diffusion mode for PDE data generation, \citealt{apte2023diffusion} reported that they generated 10,000 pairs of $[f,u]$ that satisfy equation \eqref{eq:Poisson2D} on a 64$\times$64 grid on the domain $\Omega$ using a multigrid solver. 
We decided to instead generate a dataset of analytical solutions so that our diffusion model would not learn the numerical error associated with numerical solutions.

For our training dataset, we generated 38,250 samples for the diffusion model based on analytical solutions of equation \eqref{eq:Poisson2D}. 
To do this we used (i) functions based on a neural network (appendix \ref{sec:nn_data}) and (ii) analytical solutions by choosing differentiable $u$ that satisfies the boundary condition and solving for $f$ directly (appendix \ref{app:analytical}). This data was used to train the diffusion model for the generation of PDE data.

\subsection{Diffusion model}\label{sec:diffusion_PDEgen}
We replicated the method of PDE solution generation of the 2D Poisson equation with Dirichlet boundary conditions as done in \cite{apte2023diffusion}, by training a diffusion model by modifying the GitHub repository by \citealt{vsehwagGithub} that is based on DDIM. This model involves an inverse (or ``diffusion process'') that is a Markov chain, which gradually adds Gaussian noise to the data given by a cosine scheduler.


We note that this is a large diffusion model so it is very computationally expensive to train. 
We trained our diffusion model on Compute Canada using 4 V100 GPUs for our dataset of 38,250 PDE solution samples (see Section \ref{sec:data}) and the training time was approximately 4 days.

Some of the unconditionally generated data are posted in Appendix \ref{app:gen1}. We compared generated pairs of $u$ and $f$ (labeled ``u'' and ``f'' in Figure \ref{fig:gen_figs1}) with the finite difference solution (labeled ``Finite difference solution'' in Figure \ref{fig:gen_figs1}) for comparison. Our MAE between the generated $u(x,y)$ and the finite difference solution is $4.373\times 10^{-4}$, thus showing that the generated solutions are a good approximation to the true solution.



\section{Numerical results}

We conducted our experiments on the dataset described in Section \ref{sec:data} using a trained DDIM model. Our test set is 1024 samples of neural network pairs described in Appendix \ref{sec:nn_data} with seeds separate from the training set. 
To ensure the reproducibility of our results, we posted our code in the following GitHub repository: \href{https://github.com/amartyamukherjee/minimal-diffusion}{https://github.com/amartyamukherjee/minimal-diffusion}. 
Restoring a single batch (of 1024 samples) took approximately 1 minute in a V100 GPU.

We have included results from other data-driven methods for solving PDEs\textemdash PINNs \citep{raissi2019physics} and physics-informed DeepONet \citep{lu2019deeponet}. The results from PINNs were each trained directly on the test set. They were trained in the inverse problem by learning to interpolate the grid points in $u(x,y)$, and we compute $f(x,y)$ by computing its Laplacian on the grid points using auto-differentiation. They were trained in the forward problem by setting $u=0$ as the boundary condition and adding a physics loss along the grid points of $f(x,y)$. DeepONets were tested identically but through the train set instead. They take either $u(x,y)$ or $f(x,y)$ as inputs to the branch net depending on the problem, and coordinate $(x,y)$ as the input to the trunk net.

\subsection{Sampling $u$ conditioned on $f$}

Our parameters are $\eta=8\times 10^{-9}$ and $\eta_b=9\times 10^{-9}$. We used $\sigma_f=1\times 10^{-6}$ to ensure that the sampling of $\mathbf{u}_T$ can be done for any $\sigma_T>1$. Our results are posted in table \ref{tab:mae_u}.

Unfortunately, despite being trained on the test set itself, PINNs had an average MAE of 0.002156, which is worse than the dry forward process conducted with our DDIM model. The DeepONet had an average MAE of $3.183\times 10^{-4}$, which outperforms the dry forward process conducted with our DDIM model.

Upon using our modified DDRM to sample $u(x,y)$ conditioned on $f(x,y)$, we got an MAE of $1.175\times 10^{-6}$, which is a significant improvement compared to using the DDIM model without restoration and the DeepONet. For qualitative results, we posted three samples in Appendix \ref{app:ddrm_u}. This, as a result, demonstrates the practicality of DDRM in solving problems in PDEs. 
This method, however, does not outperform finite difference approximation that had an MAE of $6.672\times 10^{-7}$ on the test set in approximating $u(x,y)$. 
\begin{table}[H]
    \centering
    \begin{tabular}{c|c|c|c|c}
        PINNs & Dry forward Process & PI-DeepONet & DDRM & Finite Difference\\
        \hline
        2.156e-03 & 1.123e-03 & 3.183e-04 & 1.175e-06 & 6.672e-07\\
    \end{tabular}
    \caption{MAE in predicting $u$ conditioned on $f$, averaged along 1024 samples}
    \label{tab:mae_u}
\end{table}

\subsection{Sampling $f$ conditioned on $u$}

Our parameters are $\eta=8\times 10^{-4}$ and $\eta_b=9\times 10^{-4}$ to ensure slow and stable incremental convergence. We used $\sigma_{n,m}=1e-6$ for all $n,m$ to ensure that the sampling of $\mathbf{f}_T$ can be done for any $\sigma_T>1$. Our results are posted in table \ref{tab:mae_f}.

The DeepONet had an average MAE of $4.224\times 10^{-1}$, which outperforms the dry inverse process conducted with our DDIM model. PINNs achieve a better MAE of $3.704\times 10^{-1}$, even though this is because it directly interacts with the test set.

Upon using our modified DDRM to sample $f(x,y)$ conditioned on $u(x,y)$, we got an MAE of $3.215\times 10^{-2}$, which is a significant improvement compared to using the DDIM model without restoration. For qualitative results, we posted three samples in Appendix \ref{app:ddrm_f}. This, as a result, demonstrates the practicality of DDRM in solving forward problems in PDEs.

This method, however, does not outperform finite difference approximation that had an MAE of $1.663\times 10^{-2}$ on the test set in approximating $f(x,y)$. This is because finite difference methods do not assume any notion of periodicity in the dataset, unlike the DST method, thus approximating the Laplacian in the boundary more reliably.
\begin{table}[H]
    \centering
    \begin{tabular}{c|c|c|c|c}
        PINNs & PI-DeepONet & Dry inverse Process & DDRM & Finite Difference\\
        \hline
        3.704e-01&4.224e-01&5.515e-01 & 3.215e-02 & 1.163e-02\\
    \end{tabular}
    \caption{MAE in predicting $f$ conditioned on $u$, averaged along 1024 samples}
    \label{tab:mae_f}
\end{table}

\section{Discussion}
In this study, we first reproduced the methods for PDE data generation in \citealt{apte2023diffusion}. The authors showed that the diffusion model could effectively generate pairs of PDE solutions for the 2D Poisson equation \eqref{eq:Poisson2D} based on both visual and statistical analyses. They reported that pairs of functions adhered to the underlying physics despite not incorporating the physics into the loss function, as would be done in PINNs \citep{raissi2019physics,blechschmidt2021three}.
Upon reproducing their work, we noticed that the diffusion model outputs a noisy estimation of the solution conditioned on the parameters, and it struggles to estimate the parameters conditioned on the solution. To solve this problem, we employed DDRM, which effectively recovers the solution and the parameters by projecting them into the shared spectral space, which in our paper, is the DST. This method outperforms other data-driven methods including PINNs and DeepONets in the forward and inverse problem.

Indeed, the dataset used for training is different from the dataset used in \citealt{apte2023diffusion} due to the lack of information about the PDEs used in their study other than the size of the grid and that they used multigrid methods to generate pairs of $[f,u]$. Our approach differs in that we chose to use data that includes differentiable $u$ that satisfies the boundary conditions and $f$ computed directly. Using a different training dataset will result in a different trained diffusion model. However, our results did show that the diffusion model was able to generate paired PDE solutions that adhere to physics laws for certain function types, as shown in Appendix \ref{app:gen1}. 

The diffusion model used is popular within this field and has shown success for image generation of various types of datasets from MNIST to celebrity faces to melanoma images \citep{vsehwagGithub}. However, in our PDE data generation example we seek to generate not just one ``image'' or function solution, but pairs of functions $[f,u]$ that adhere to the physics of the PDE, which in this case is the Poisson equation \eqref{eq:Poisson2D}. This is what makes this approach novel and more difficult than previous applications. 




\subsection{Future directions}

In this study, we initiated our exploration of diffusion models for solving PDEs, starting with the Poisson equation. The Poisson equation's analytical solvability for a wide range of differentiable functions $u$ and the well-established nature of finite difference methods for it provide a robust baseline against which we can evaluate the performance of our diffusion model. 
Our results, which outperform other data-driven methods, suggest that the methodologies we have developed and validated for the Poisson equation can serve as a foundational blueprint for exploring solutions to higher-order PDEs, such as the Navier-Stokes equation.



Diffusion models are a new and very active area of research. While the diffusion models used in our paper do not directly include physics in their training process, future work may be able to include physics more directly in a similar approach to physics-informed neural networks \citep{raissi2019physics,blechschmidt2021three, zhang_physics-informed_2022}. We do note that diffusion models do seem to adhere to physics after training as shown in \citealt{apte2023diffusion}, but still struggle to outperform finite difference methods.


\acks{We sincerely acknowledge the support of Professor Matthew Scott from the Department of Applied Mathematics, University of Waterloo, and Professor Yaoliang Yu from the Department of Computer Science, University of Waterloo for their valuable feedback on the paper.}

\bibliography{colt2024}

\newpage
\appendix


\section{Proof of Proposition \ref{thm:eigenpow}}\label{app:pf_eigenpow}

\begin{proof}
    To find the eigenvalues and eigenfunctions of the Laplacian operator, we are essentially solving the following PDE
    \begin{equation}
        \Delta u=\lambda u.
    \end{equation}
    We use separation of parts to split $u$ as $u(x,y)=X(x)Y(y)$. It follows that 
    \begin{equation}
        \Delta u=X''(x)Y(y)+X(x)Y''(y)=\lambda X(x)Y(y),
    \end{equation}
    \begin{equation}
        \frac{X''}{X}+\frac{Y''}{Y}=\lambda.
    \end{equation} 
    Here, we solve two eigenvalue problems, $X''/X=\lambda_1$ and $Y''/Y=\lambda_2$, which means $\lambda=\lambda_1+\lambda_2$. Consider the boundary value problem for $X$: 
    \begin{equation}
        \frac{X''}{X}=\lambda_1,X(0)=X(1)=0. 
    \end{equation}
    The eigenvalue and eigenfunction pairs are trivially of the form
    \begin{equation}
        X_n(x)=\sin(n\pi x),\lambda_{1,n}=-(n\pi)^2.
    \end{equation}
    By symmetry, we have
    \begin{equation}
        Y_m(y)=\sin(m\pi y),\lambda_{2,m}=-(m\pi)^2.
    \end{equation}
    Thus, the eigenvalue and eigenfunction pairs of the Laplacian operator are of the form
    \begin{align}
        &u_{n,m}(x,y)=\sin(n\pi x)\sin(m\pi y),\\&\lambda_{n,m}=-(n\pi)^2-(m\pi)^2.
    \end{align}
\end{proof}

\section{Proof of Theorem \ref{thm:greens_distribution}}\label{app:pf_greens_distribution}

\begin{proof}
In a bounded domain $\Omega\subset\R^2$ with smooth boundary, any function $u\in C^2(\overline\Omega)$ satisfies
\begin{align}
    u(x,y)
    =&\iint_\Omega\psi((x',y')-(x,y))\Delta u(x',y')dx'dy'\nonumber\\
    &+\int_{\partial\Omega}u(x',y')\frac{\partial\psi}{\partial n}((x',y')-(x,y))dS_{x',y'}\nonumber\\
    &-\int_{\partial\Omega}\psi((x',y')-(x,y))\frac{\partial u}{\partial n}(x',y')dS_{x',y'},\label{eq:gilbarg}
\end{align}
where $\psi(x,y)=\frac{\ln(||(x,y)||}{2\pi}$ is Green's function in two dimensions. We commonly refer to the first integral as the Newtonian potential, the second integral as the double-layer potential, and the third integral as the single-layer potential (we will use these terms in the proof of Theorem \ref{thm:dst_u_var} in Appendix \ref{app:pf_dst_u_var}). This known result was retrieved from \cite{gilbarg1983elliptic}. Given that $\Delta u=f+z$ with $z\sim \mathcal{N}(0,\sigma_f^2)$, the first integrand can be written as
\begin{align}
    &\iint_\Omega\psi((x',y')-(x,y))\Delta u(x',y')dx'dy'\nonumber\\
    =&\iint_\Omega\psi((x',y')-(x,y))fdx'dy'\nonumber\\
    &+\iint_\Omega\psi((x',y')-(x,y))zdx'dy'.\label{eq:greens_dist_stochastic}
\end{align}
Consequently, it follows that $E[u(x,y)]=u_0$, where $u_0$ is the deterministic solution to $\Delta u_0=f$. Furthermore, we can compute $Var[u(x,y)]=\sigma_f^2K(x,y)$, where 
\begin{equation}
    K(x,y)=\iint_{\Omega}(\psi((x',y')-(x,y)))^2dx'dy'.
\end{equation}
\end{proof}

\section{Proof of Theorem \ref{thm:dst_u_var}}\label{app:pf_dst_u_var}

\begin{proof}
This proof follows from the proof of Theorem \ref{thm:greens_distribution} provided in Appendix \ref{app:pf_greens_distribution}. Consider the stochastic component of equation \eqref{eq:greens_dist_stochastic}: 
\begin{equation}
    u_z(x,y)=\iint_\Omega\psi((x',y')-(x,y))zdx'dy',
\end{equation}
where $z\sim N(0,\sigma_f^2)$. Let us compute the discrete sine transformation (DST) of $u_z(x,y)$: 
\begin{align}
    \overline{\mathbf{u}_z}^{(n,m)}
    &=\iint_\Omega\iint_\Omega\psi((x',y')-(x,y))zdx'dy'\sin(n\pi x)\sin(m\pi y)dxdy\\
    &=\iint_\Omega\left[\iint_\Omega\psi((x',y')-(x,y))\sin(n\pi x)\sin(m\pi y)dxdy\right]zdx'dy'. 
\end{align}
Notice that the term in the square brackets is the DST of $\psi((x',y')-(x,y))$. It is also identical to the Newtonian potential from equation \eqref{eq:gilbarg}. This is identical to computing an analytical solution for the following PDE:
\begin{equation}
    \Delta u=\sin(nx)\sin(mx),~u=0\text{ in }\partial\Omega,
\end{equation}
where the known solution is
\begin{equation}
    u=-\frac{1}{\pi^2(n^2+m^2)}\sin(nx)\sin(mx).
\end{equation}
Plugging this into equation \eqref{eq:gilbarg} gives 
\begin{align}
    -\frac{1}{\pi^2(n^2+m^2)}\sin(nx)\sin(mx)
    =&\iint_\Omega\psi((x',y')-(x,y))\sin(n\pi x')\sin(m\pi y')dx'dy'\nonumber\\
    &+\int_{\partial\Omega}u(x',y')\frac{\partial\psi}{\partial n}((x',y')-(x,y))dS_{x',y'}\nonumber\\
    &-\int_{\partial\Omega}\psi((x',y')-(x,y))\frac{\partial u}{\partial n}(x',y')dS_{x',y'}.
\end{align}
Due to the boundary conditions, the double-layer potential term is 0. We can compute the gradient vector of $u$:
\begin{equation}
    \nabla u(x,y)=\begin{pmatrix}
        n\pi\cos(n\pi x)\sin(m\pi y)\\
        m\pi\sin(n\pi x)\cos(m\pi y)
    \end{pmatrix}.
\end{equation}
The single-layer potential is expensive to compute, but we can bound it.
\begin{align}
    &|\int_{\partial\Omega}\psi((x',y')-(x,y))\frac{\partial u}{\partial n}(x',y')dS_{x',y'}|\\
    &\leq\sup_{(x',y')}|\psi((x',y')-(x,y))|\|\nabla u(x,y)\|\int_{\partial\Omega}dS_{x',y'}.
\end{align}
Since $\Omega=[0,1]^2$, $\int_{\partial\Omega}dS_{x',y'}$ is the sum of its edges, which is $4$. The maximum of $\nabla u$ along the boundary is $\pi\max(n,m)$. The supremum of $|\psi((x',y')-(x,y))|$ is
\begin{equation}
    \sup_{(x',y')}|\psi((x',y')-(x,y))|=|\psi((1,1)-(0,0))|=\frac{\ln\sqrt{2}}{2\pi}=\frac{\ln 2}{4\pi}.
\end{equation}
Putting it all together, we get the following upper-bound
\begin{align}
    &|\int_{\partial\Omega}\psi((x',y')-(x,y))\frac{\partial u}{\partial n}(x',y')dS_{x',y'}|\\
    &\leq\frac{\ln 2}{4\pi}\cdot\pi\max(n,m)\cdot 4\\
    &=\ln 2\max(n,m),
\end{align}
which gives us the following upper bound for the variance
\begin{align}
    Var[\overline{\mathbf{u}_z}^{(n,m)}|f_0]
    &=Var\left[\iint_\Omega\left[\iint_\Omega\psi((x',y')-(x,y))\sin(n\pi x)\sin(m\pi y)dxdy\right]zdx'dy'\right]\\
    &\leq Var\left[\iint_\Omega\left[\frac{1}{\pi^2(n^2+m^2)}|\sin(nx')\sin(mx')|+\ln 2\max(n,m)\right]zdx'dy'\right]\\
    &\leq Var\left[\iint_\Omega\left[\frac{1}{\pi^2(n^2+m^2)}+\ln 2\max(n,m)\right]zdx'dy'\right]\\
    &=(\frac{1}{\pi^2(n^2+m^2)}+\ln 2\max(n,m))^2\sigma_f^2.
\end{align}
\end{proof}

\section{Proof of Proposition \ref{prop:variational_inverse}}
\label{app:pf_variational_inverse}

\begin{proof}
This proof uses properties of Gaussian marginals. We refer the reader to \cite{bishop2006pattern}.

\subsection*{Distribution of $\overline{\mathbf{f}}_T^{(n,m)}$}

We have equation \eqref{eq:variational_inverse_1}.
\begin{equation}
    q^{(T)}(\overline{\mathbf{f}}_T^{(n,m)} \mid {\mathbf{f}}_0, \mathbf{u})=\mathcal{N}(\overline{\mathbf{u}}^{(n,m)},\sigma_T^2-\sigma_{n,m}^2\lambda_{n,m}^2).
\end{equation}
We can assume that $q^{(T)}(\overline{\mathbf{f}}_T^{(n,m)} \mid {\mathbf{f}}_0, \mathbf{u})=q^{(T)}(\overline{\mathbf{f}}_T^{(n,m)} \mid \overline{\mathbf{f}}_0^{(n,m)},\overline{\mathbf{u}}^{(n,m)})$,
and we know from Section \ref{sec:condition_dist_inverse} that
\begin{equation}
    p(\overline{\mathbf{f}}_0^{(n,m)} \mid \overline{\mathbf{u}}^{(n,m)})=\mathcal{N}(\overline{\mathbf{f}}_0^{(n,m)},\sigma_{n,m}^2\lambda_{n,m}^2).
\end{equation}
Using the property of Gaussian marginals, we can derive the following result
\begin{equation}
    q^{(T)}(\overline{\mathbf{f}}_T^{(n,m)} \mid {\mathbf{f}}_0)=\mathcal{N}(\overline{\mathbf{f}}_0^{(n,m)},\sigma_{n,m}^2\lambda_{n,m}^2+\sigma_T^2-\sigma_{n,m}^2\lambda_{n,m}^2)=\mathcal{N}(\overline{\mathbf{f}}_0^{(n,m)},\sigma_T^2).
\end{equation}

\subsection*{Distribution of $\overline{\mathbf{f}}_t^{(n,m)}$}

We have equation \eqref{eq:variational_inverse_t}. Consider the condition where $\sigma_t<\sigma_{n,m}{\lambda_{n,m}}$.
\begin{equation}
    q^{(t)}\left(\overline{\mathbf{f}}_t^{(n,m)} \mid {\mathbf{f}}_0,\mathbf{f}_{t+1}, \mathbf{u}\right)=\mathcal{N}\Big(\overline{\mathbf{f}}_{0}^{(n,m)}+\sqrt{1-\eta^2} \sigma_t \frac{\overline{\mathbf{u}}^{(n,m)}-\overline{\mathbf{f}}_{0}^{(n,m)}}{\sigma_{n,m} \lambda_{n,m}},\eta^2 \sigma_t^2\Big).
\end{equation}
We can similarly assume that $q^{(T)}(\overline{\mathbf{f}}_T^{(n,m)} \mid {\mathbf{f}}_0, \mathbf{f}_{t+1}, \mathbf{u})=q^{(T)}(\overline{\mathbf{f}}_T^{(n,m)} \mid \overline{\mathbf{f}}_0^{(n,m)},\overline{\mathbf{f}}_{t+1}^{(n,m)}, \overline{\mathbf{u}}^{(n,m)})$. And we can also state that $q^{(T)}(\overline{\mathbf{f}}_T^{(n,m)} \mid \overline{\mathbf{f}}_0^{(n,m)},\overline{\mathbf{f}}_{t+1}^{(n,m)}, \overline{\mathbf{u}}^{(n,m)})=q^{(T)}(\overline{\mathbf{f}}_T^{(n,m)} \mid \overline{\mathbf{f}}_0^{(n,m)},\overline{\mathbf{u}}^{(n,m)})$ since there is no dependence on $\overline{\mathbf{f}}_{t+1}^{(n,m)}$ in the distribution. We use the property that $\frac{\overline{\mathbf{u}}^{(n,m)}-\overline{\mathbf{f}}_{0}^{(n,m)}}{\sigma_{n,m} \lambda_{n,m}}$ is a standard Gaussian. Using the property of Gaussian marginals, we can derive the following result
\begin{align}
    q^{(t)}\left(\overline{\mathbf{f}}_t^{(n,m)} \mid {\mathbf{f}}_0\right)=\mathcal{N}\Big(&\overline{\mathbf{f}}_{0}^{(n,m)}+\sqrt{1-\eta^2} \sigma_t \frac{\overline{\mathbf{u}}^{(n,m)}-\overline{\mathbf{f}}_{0}^{(n,m)}}{\sigma_{n,m} \lambda_{n,m}}-\sqrt{1-\eta^2} \sigma_t \frac{\overline{\mathbf{u}}^{(n,m)}-\overline{\mathbf{f}}_{0}^{(n,m)}}{\sigma_{n,m} \lambda_{n,m}},\nonumber\\&\eta^2 \sigma_t^2+(1-\eta^2)\sigma_t^2\Big)\\
    =\mathcal{N}(&\overline{\mathbf{f}}_0^{(n,m)},\sigma_t^2).
\end{align}

Next, consider the condition where $\sigma_t\geq\sigma_{n,m}{\lambda_{n,m}}$.
\begin{equation}
    q^{(t)}\left(\overline{\mathbf{f}}_t^{(n,m)} \mid {\mathbf{f}}_0,\mathbf{f}_{t+1}, \mathbf{u}\right)=\mathcal{N}\Big(\left(1-\eta_b\right) \overline{\mathbf{f}}_{0}^{(n,m)}+\eta_b \overline{\mathbf{u}}^{(n,m)},\sigma_t^2-{\sigma_{n,m}^2}{\lambda_{m,n}^2} \eta_b^2\Big)
\end{equation}
Using the property that $\eta_b(\overline{\mathbf{f}}_{0}^{(n,m)}+\overline{\mathbf{u}}^{(n,m)})=\mathcal{N}(0,{\sigma_{n,m}^2}{\lambda_{m,n}^2} \eta_b^2)$, we can derive using Gaussian marginals
\begin{align}
    q^{(t)}\left(\overline{\mathbf{f}}_t^{(n,m)} \mid {\mathbf{f}}_0\right)=\mathcal{N}\Big(&\left(1-\eta_b\right) \overline{\mathbf{f}}_{0}^{(n,m)}+\eta_b \overline{\mathbf{u}}^{(n,m)}+\eta_b(\overline{\mathbf{f}}_{0}^{(n,m)}+\overline{\mathbf{u}}^{(n,m)}),\nonumber\\&\sigma_t^2-{\sigma_{n,m}^2}{\lambda_{m,n}^2} \eta_b^2+{\sigma_{n,m}^2}{\lambda_{m,n}^2} \eta_b^2\Big)\\
    =\mathcal{N}(&\overline{\mathbf{f}}_0^{(n,m)},\sigma_t^2).
\end{align}
\end{proof}

\section{Proof of Proposition \ref{prop:variational_forward}}
\label{app:pf_variational_forward}

\begin{proof}
We notify the reader that the proof of this proposition is nearly a repetition of the proof of Proposition \ref{prop:variational_inverse} presented in Appendix \ref{app:pf_variational_inverse}, with the only difference being the variance terms in the inverse and forward variational distributions.

This proof uses properties of Gaussian marginals. We refer the reader to \cite{bishop2006pattern}.

\subsection*{Distribution of $\overline{\mathbf{u}}_T^{(n,m)}$}

We have equation \eqref{eq:variational_forward_1}.
\begin{equation}
    q^{(T)}(\overline{\mathbf{u}}_t^{(n,m)} \mid \mathbf{u}_0,\mathbf{f})=\mathcal{N}(\overline{\mathbf{f}}^{(n,m)},\sigma_T^2-\sigma_{f}^2\overline{\mathbf{K}}^{(n,m)}/\lambda_{n,m}^2),
\end{equation}
We can assume that $q^{(T)}(\overline{\mathbf{u}}_T^{(n,m)} \mid {\mathbf{u}}_0, \mathbf{f})=q^{(T)}(\overline{\mathbf{u}}_T^{(n,m)} \mid \overline{\mathbf{u}}_0^{(n,m)},\overline{\mathbf{f}}^{(n,m)})$,
and we know from Section \ref{sec:condition_dist_forward} that
\begin{equation}
    p(\overline{\mathbf{u}}_0^{(n,m)} \mid \overline{\mathbf{f}}^{(n,m)})=\mathcal{N}(\overline{\mathbf{u}}_0^{(n,m)},\sigma_{f}^2\overline{\mathbf{K}}^{(n,m)}/\lambda_{n,m}^2).
\end{equation}
Using the property of Gaussian marginals, we can derive the following result
\begin{equation}
    q^{(T)}(\overline{\mathbf{u}}_t^{(n,m)} \mid \mathbf{u}_0)=\mathcal{N}\left(\overline{\mathbf{u}}_0^{(n,m)},\sigma_T^2-\sigma_{f}^2\overline{\mathbf{K}}^{(n,m)}/\lambda_{n,m}^2+\sigma_{f}^2\overline{\mathbf{K}}^{(n,m)}/\lambda_{n,m}^2\right)=\mathcal{N}\left(\overline{\mathbf{u}}_0^{(n,m)},\sigma_T^2\right).
\end{equation}

\subsection*{Distribution of $\overline{\mathbf{u}}_t^{(n,m)}$}

We have equation \eqref{eq:variational_forward_t}. Consider the condition where $\sigma_t<\frac{\sigma_{f}\sqrt{\overline{\mathbf{K}}^{(n,m)}}}{\lambda_{n,m}}$
\begin{equation}
    q^{(t)}\left(\overline{\mathbf{u}}_t^{(n,m)} \mid \mathbf{u}_0,\mathbf{u}_{t+1}, \mathbf{f}\right) = \mathcal{N}\Big(\overline{\mathbf{u}}_{0}^{(n,m)}+\sqrt{1-\eta^2} \sigma_t \frac{\overline{\mathbf{f}}^{(n,m)}-\overline{\mathbf{u}}_{0}^{(n,m)}}{\sigma_{f}\sqrt{\overline{\mathbf{K}}^{(n,m)}} / \lambda_{n,m}},\eta^2 \sigma_t^2\Big)
\end{equation}
We can similarly assume that $q^{(T)}(\overline{\mathbf{u}}_T^{(n,m)} \mid {\mathbf{u}}_0, \mathbf{u}_{t+1}, \mathbf{f})=q^{(T)}(\overline{\mathbf{u}}_T^{(n,m)} \mid \overline{\mathbf{u}}_0^{(n,m)},\overline{\mathbf{u}}_{t+1}^{(n,m)}, \overline{\mathbf{f}}^{(n,m)})$. And we can also state that $q^{(T)}(\overline{\mathbf{u}}_T^{(n,m)} \mid \overline{\mathbf{u}}_0^{(n,m)},\overline{\mathbf{u}}_{t+1}^{(n,m)}, \overline{\mathbf{f}}^{(n,m)})=q^{(T)}(\overline{\mathbf{u}}_T^{(n,m)} \mid \overline{\mathbf{u}}_0^{(n,m)},\overline{\mathbf{f}}^{(n,m)})$ since there is no dependence on $\overline{\mathbf{u}}_{t+1}^{(n,m)}$ in the distribution. We use the property that $(\overline{\mathbf{f}}^{(n,m)}-\overline{\mathbf{u}}_{0}^{(n,m)})/\left(\sigma_{f}\sqrt{\overline{\mathbf{K}}^{(n,m)}}/\lambda_{n,m}\right)$ is a standard Gaussian. Using the property of Gaussian marginals, we can derive the following result
\begin{align}
    q^{(t)}\left(\overline{\mathbf{u}}_t^{(n,m)} \mid \mathbf{u}_0\right) = \mathcal{N}\Big(&\overline{\mathbf{u}}_{0}^{(n,m)}+\sqrt{1-\eta^2} \sigma_t \frac{\overline{\mathbf{f}}^{(n,m)}-\overline{\mathbf{u}}_{0}^{(n,m)}}{\sigma_{f}\sqrt{\overline{\mathbf{K}}^{(n,m)}} / \lambda_{n,m}}\nonumber\\&-\sqrt{1-\eta^2} \sigma_t \frac{\overline{\mathbf{f}}^{(n,m)}-\overline{\mathbf{u}}_{0}^{(n,m)}}{\sigma_{f}\sqrt{\overline{\mathbf{K}}^{(n,m)}} / \lambda_{n,m}},\nonumber\\&\eta^2 \sigma_t^2+(1-\eta^2) \sigma_t^2\Big)\\
    =\mathcal{N}\Big(&\overline{\mathbf{u}}_{0}^{(n,m)},\sigma_t^2\Big)
\end{align}

Nwext, consider the condition where $\sigma_t\geq\frac{\sigma_{f}\sqrt{\overline{\mathbf{K}}^{(n,m)}}}{\lambda_{n,m}}$
\begin{equation}
    q^{(t)}\left(\overline{\mathbf{u}}_t^{(n,m)} \mid \mathbf{u}_0,\mathbf{u}_{t+1}, \mathbf{f}\right) = \mathcal{N}\Big(\left(1-\eta_b\right) \overline{\mathbf{u}}_{0}^{(n,m)}+\eta_b \overline{\mathbf{f}}^{(n,m)},\sigma_t^2-\frac{\sigma_{f}^2\overline{\mathbf{K}}^{(n,m)}}{\lambda_{n,m}^2} \eta_b^2\Big)
\end{equation}
Using the property that $\eta_b(\overline{\mathbf{u}}_{0}^{(n,m)}+\overline{\mathbf{f}}^{(n,m)})=\mathcal{N}(0,\frac{\sigma_{f}^2\overline{\mathbf{K}}^{(n,m)}}{\lambda_{n,m}^2} \eta_b^2)$, we can derive using Gaussian marginals
\begin{align}
    q^{(t)}\left(\overline{\mathbf{u}}_t^{(n,m)} \mid \mathbf{u}_0\right) = \mathcal{N}\Big(&\left(1-\eta_b\right) \overline{\mathbf{u}}_{0}^{(n,m)}+\eta_b \overline{\mathbf{f}}^{(n,m)}+\eta_b(\overline{\mathbf{u}}_{0}^{(n,m)}+\overline{\mathbf{f}}^{(n,m)}),\nonumber\\&\sigma_t^2-\frac{\sigma_{f}^2\overline{\mathbf{K}}^{(n,m)}}{\lambda_{n,m}^2} \eta_b^2+\frac{\sigma_{f}^2\overline{\mathbf{K}}^{(n,m)}}{\lambda_{n,m}^2} \eta_b^2\Big)\\
    =\mathcal{N}\Big(&\overline{\mathbf{u}}_{0}^{(n,m)},\sigma_t^2\Big)
\end{align}
\end{proof}

\section{Dataset solutions}\label{app:dataset}
We generated 38,250 samples for the diffusion model for PDE data generation of the following types: neural network pairs and analytical pairs. Here we show some examples of these different types of data.

\subsection{Neural network pairs}
\label{sec:nn_data}
\begin{figure}[H]
    \centering
     \subfigure{\includegraphics[width=0.7\linewidth,trim={2cm 3cm 2cm 3cm},clip]{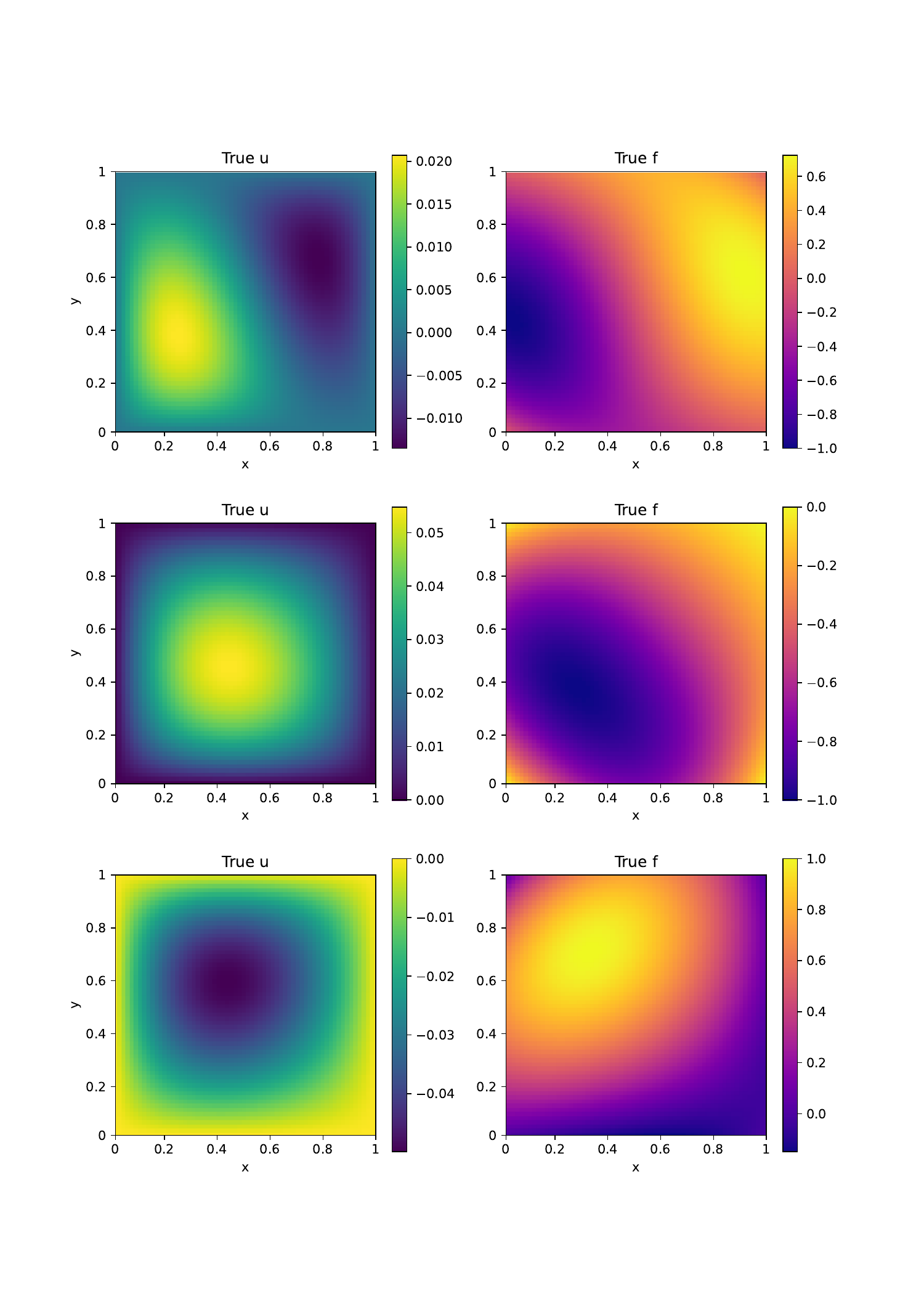}}
        \caption{Examples of analytical neural network pairs.}\label{fig:ana_type7}
\end{figure}

To respect the Dirichlet boundary conditions, we sampled $u(x,y)$ randomly as:
\begin{equation}
    u(x,y)=g_{NN}(x,y)x(1-x)y(1-y),
\end{equation}
where $g_{NN}$ is a randomly initialized neural network with three hidden layers and \textit{tanh} activation function.
Clearly $u$ satisfies the boundary conditions since when $x$ or $y$ is 1 or 0, $u = 0$. Additionally, $u$ is differentiable because of the use of the \textit{tanh} activation function. We then compute $f=\Delta u$ using auto-differentiation by using the {\tt autograd.grad} function from {\tt Pytorch}. 
Using this, we generated 10,000 samples in our dataset. 
Example neural network pairs are shown in Figures \ref{fig:ana_type7}.

\subsection{Analytical pairs}\label{app:analytical}
To train a diffusion model to generate data for the 2D Poisson equation with homogeneous Dirichlet boundary condition (equation \eqref{eq:Poisson2D}) we generated training data by using smooth functions $u$ that satisfy the boundary conditions and solving for $f$ analytically. We created a $64 \times 64$ mesh grid for the domain $\Omega = [0,1]^2$ and used the analytical solution for $u$ and $f$. We classified these function pairs $[f,u]$ as different types.
\subsubsection*{Type 1 analytical pairs}
\begin{figure}[H]
    \centering
    \includegraphics[width=0.9\textwidth]{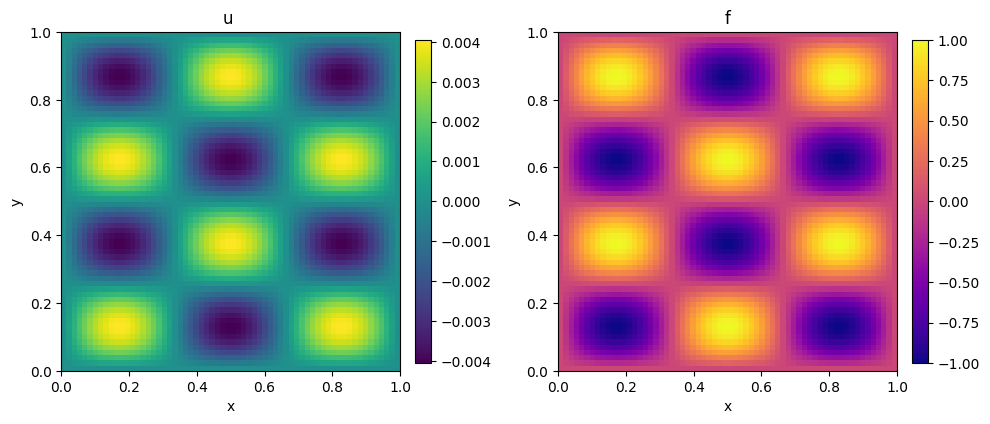}
    \caption{Example type 1 analytical solution with $n = 3$ and $k = 4$.}
    \label{fig:ana_type1}
\end{figure}

\begin{equation*}
    u(x,y) = \sin(n \pi x) \sin(k \pi y)
\end{equation*}
where $n, k$ are positive integers.
We can solve $u$ analytically to get 
\begin{equation*}
    \nabla^2 u = f = -\pi^2 (n^2 + k^2) \sin (n \pi x) \sin(k \pi y).
\end{equation*}
An example solution is shown in Fig.~\ref{fig:ana_type1}.

\subsubsection*{Type 2 analytical pairs}
\begin{figure}[H]
    \centering
    \includegraphics[width=0.9\textwidth]{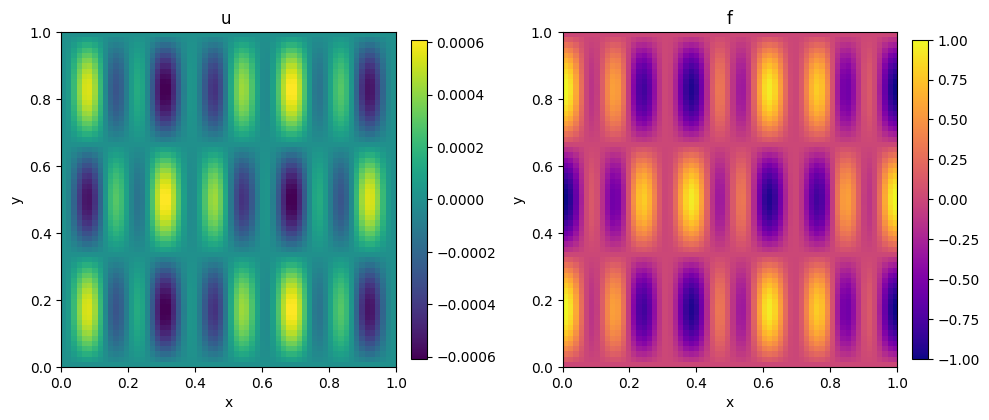}
    \caption{Example type 2 analytical solution with $n = 5$, $k = 3$, and $j = 8$.}
    \label{fig:ana_type2}
\end{figure}

\begin{equation*}
    u(x,y) = \sin(n \pi x) \sin(k \pi y) \sin(j \pi x)
\end{equation*}
for $n, k, j$ positive integers.
The analytical solution for $f$ is given by
\begin{equation*}
    \nabla^2 u = f = -\pi^2(-2 j n \cos(j\pi x) \cos(n \pi x) + (j^2 + k^2 + n^2)\sin(j \pi x) \sin(n \pi x))\sin(k \pi y).
\end{equation*}
An example solution is shown in Fig.~\ref{fig:ana_type2}.

\subsubsection*{Type 3 analytical pairs}
\begin{figure}[H]
    \centering
    \includegraphics[width=0.9\textwidth]{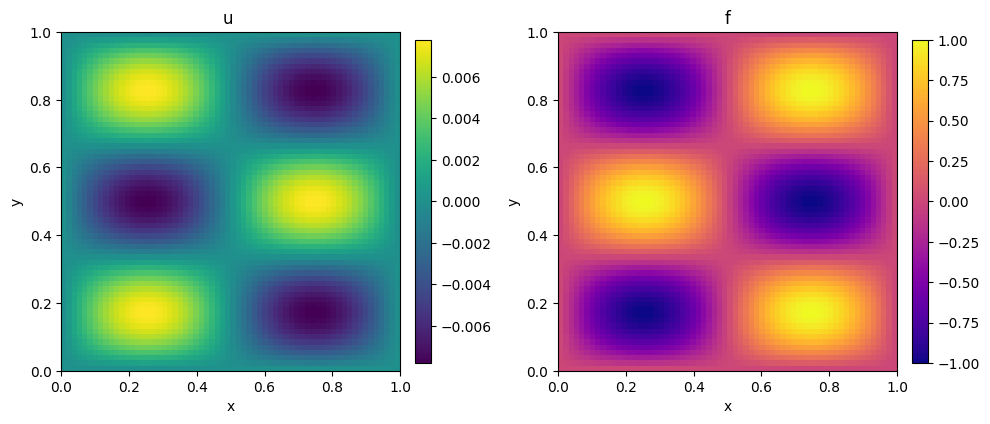}
    \caption{Example type 3 analytical solution with $n = 1$ and $k = 3$.}
    \label{fig:ana_type3}
\end{figure}

\begin{equation*}
    u(x,y) = \sin(n \pi x) \sin(k \pi y) \cos(n \pi x)
\end{equation*}
for $n, k$ positive integers. The analytical solution $f$ is given by
\begin{equation*}
    \nabla^2 u = f = -\frac{1}{2}(k^2 + 4n^2)\pi^2 \sin(2n \pi x) \sin(k \pi y).
\end{equation*}
An example solution is shown in Fig.~\ref{fig:ana_type3}.

\subsubsection*{Type 4 analytical pairs}
\begin{figure}[H]
    \centering
    \includegraphics[width=0.9\textwidth]{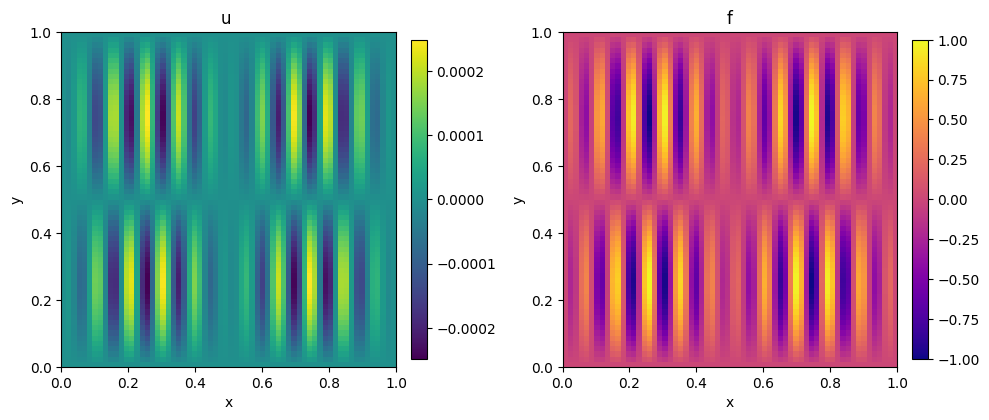}
    \caption{Example type 4 analytical solution with $n = 2$ and $k = 2$.}
    \label{fig:ana_type4}
\end{figure}

\begin{equation*}
    u(x,y) = \sin(n \pi x) \sin(k \pi y) \cos(j \pi x)
\end{equation*}
for $n, k, j$ positive integers. The analytical solution $f$ is given by
\begin{equation*}
    \nabla^2 u = f = -\pi^2(2jn \cos(n\pi x)\sin(j \pi x) + (j^2 + k^2 + n^2)(\cos(j \pi x) \sin(n \pi x)) \sin (k \pi y).
\end{equation*}
An example solution is shown in Fig.~\ref{fig:ana_type4}.
\subsubsection*{Type 5 analytical pairs}
\begin{figure}[H]
    \centering
    \includegraphics[width=0.9\textwidth]{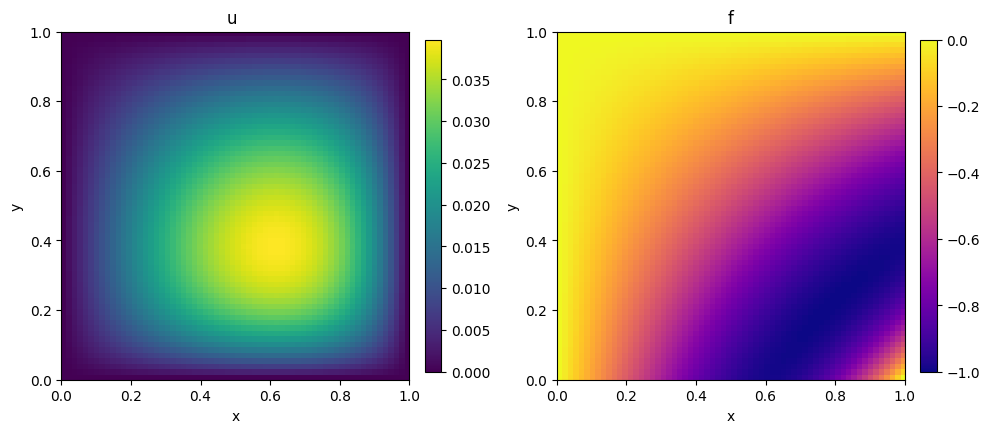}
    \caption{Example type 5 analytical solution with $n = 2$.}
    \label{fig:ana_type5}
\end{figure}

\begin{equation*}
    u(x,y) = n(x-1)x(y-1)y \exp(x - y)
\end{equation*}
for $n$ positive integers. The analytical solution $f$ is given by
\begin{equation*}
    \nabla^2 u = f = 2 \exp(x-y)nx(y-1)(2 + x(y-2) + y)
\end{equation*}
An example solution is shown in Fig.~\ref{fig:ana_type5}.
\subsubsection*{Type 6 analytical pairs}\label{app:type6}
\begin{figure}[H]
    \centering
    \includegraphics[width=0.9\textwidth]{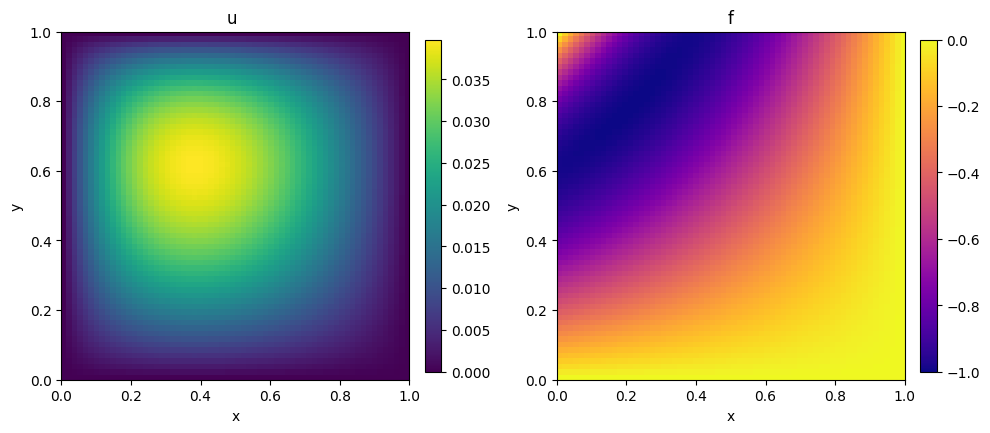}
    \caption{Example type 6 analytical solution with $n = 8$.}
    \label{fig:ana_type6}
\end{figure}

\begin{equation*}
    u(x,y) = n(x-1)x(y-1)y\exp(y-x)
\end{equation*}
for $n$ positive integers. The analytical solution $f$ is given by
\begin{equation*}
    \nabla^2 u = f = 2 n\exp(y-x) y (x-1)(2 + x - 2y + xy).
\end{equation*}
An example solution is shown in Fig.~\ref{fig:ana_type6}.

\newpage

\section{Qualitative results - Dry forward Process}\label{app:inverse_u}

\begin{figure}[H]
    \centering
    \includegraphics[width=\linewidth,trim={3cm 3cm 3cm 4cm},clip]{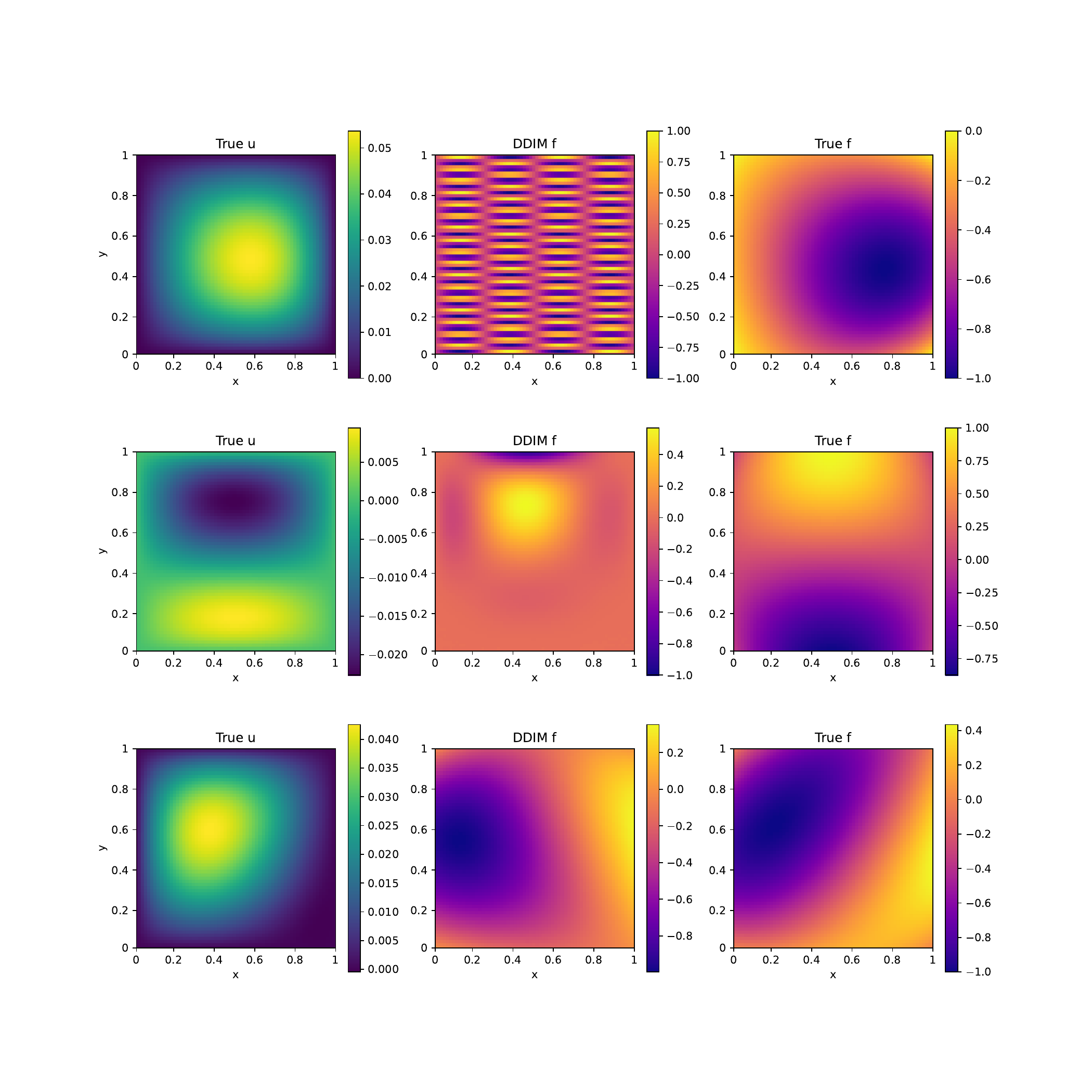}
    \caption{Plots of dry generated solutions of the inverse process $f(x,y)$ (middle), true solution $f(x,y)$ (right), and $u(x,y)$ (left).}\label{fig:inverse_u}
\end{figure}

We trained a DDIM on the dataset explained in Section \ref{sec:data} and Appendix \ref{app:dataset}. In this section, we estimate solutions to the Poisson equation $u(x,y)$ while keeping the $f(x,y)$ channel fixed. We tested this on different randomly generated neural network functions explained in Section \ref{sec:nn_data}, three of which are plotted in Figure \ref{fig:inverse_u}.

The plots show that the DDIM produces a good approximation of the solution to the Poisson equation, with some additive noise. The average MAE over 1024 different test samples is $1.123\times 10^{-3}$.


\begin{figure}[H]
    \centering
    \includegraphics[width=\linewidth,trim={3cm 3cm 3cm 4cm},clip]{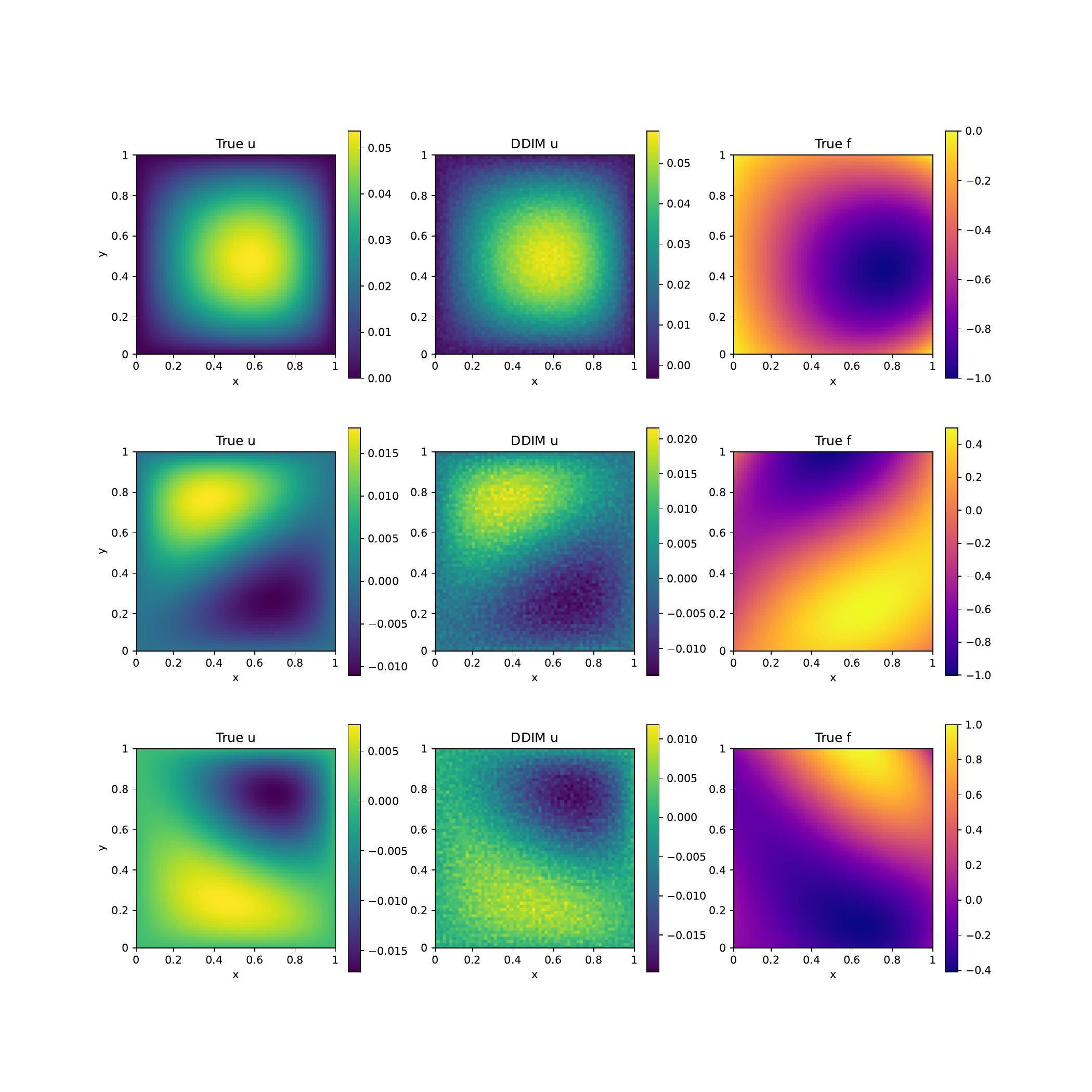}
    \caption{Plots of dry generated solutions of the forward process $u(x,y)$ (middle), $f(x,y)$ (right), and the true $u(x,y)$ (left).}\label{fig:forward_f}
\end{figure}

\section{Qualitative Results - Dry inverse Process}\label{app:forward_f}

In this section, we estimate the parameter $f(x,y)$ while keeping the $u(x,y)$ channel fixed. We tested this on different randomly generated neural network functions explained in Section \ref{sec:nn_data}, three of which are plotted in Figure \ref{fig:forward_f}.

The plots show that the DDIM produces a poor approximation of the solution to the Poisson equation. The average MAE over 1024 different test samples is $0.5515$.

\section{Qualitative Results - DDRM forward Process}\label{app:ddrm_u}

\begin{figure}[H]
    \centering
    \includegraphics[width=\linewidth,trim={3cm 3cm 3cm 4cm},clip]{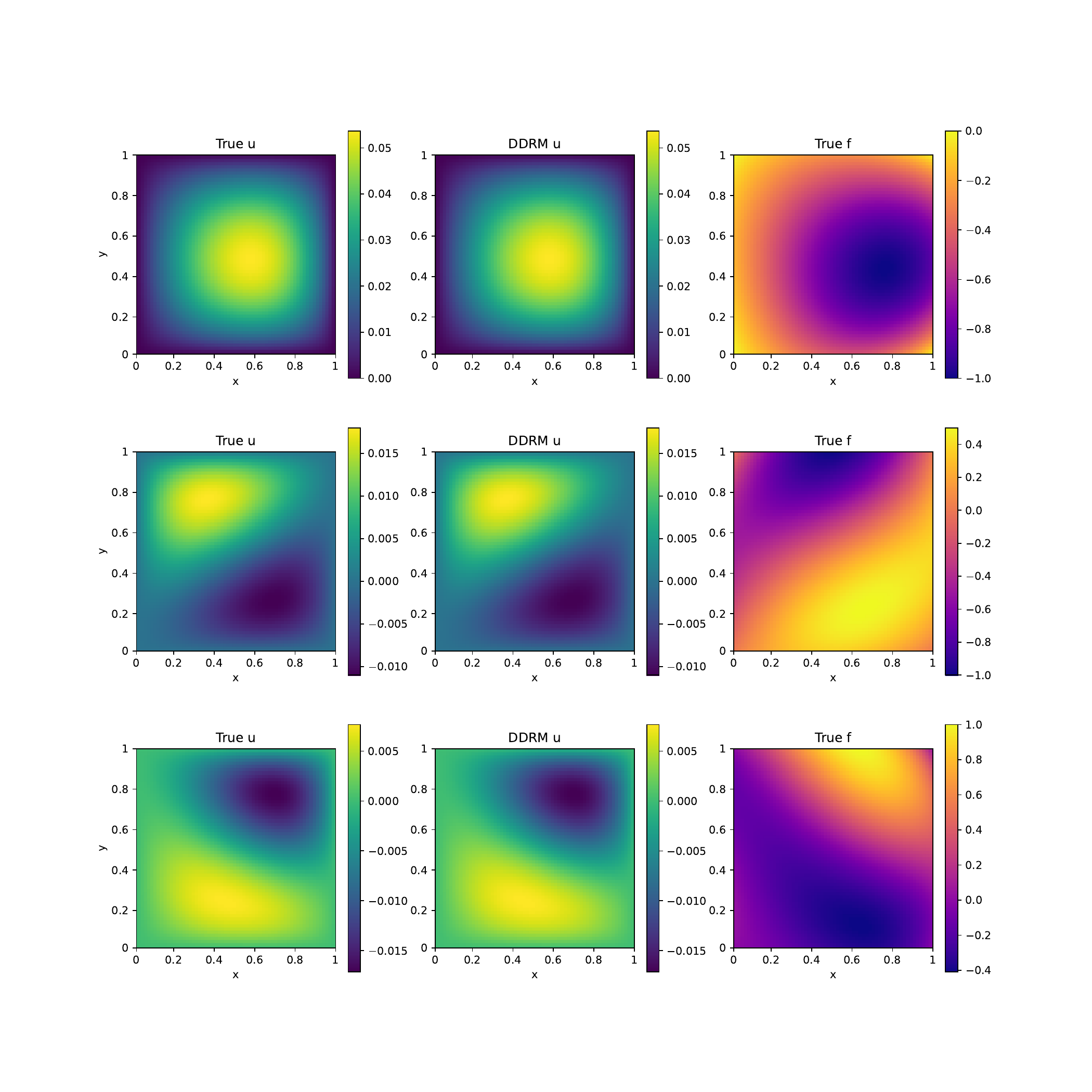}
        \caption{Plots of DDRM generated solutions of the forward process $u(x,y)$ (middle), true $u(x,y)$ (left), and $f(x,y)$ (right).}\label{fig:ddrm_u}
\end{figure}

In this section, we estimate the solution $u(x,y)$ while keeping the parameter channel $f(x,y)$ fixed using DDRM. We tested this on different randomly generated neural network functions explained in Section \ref{sec:nn_data}, three of which are plotted in Figure \ref{fig:ddrm_u}.

The plots show that the DDRM produces a significantly better approximation of the solution to the Poisson equation. The average MAE over 1024 different test samples is $1.175\times 10^{-6}$, which is just slightly greater than the MAE of $6.672\times 10^{-7}$ upon using the finite difference method.

\section{Qualitative Results - DDRM inverse Process}\label{app:ddrm_f}

\begin{figure}[H]
    \centering
    \includegraphics[width=\linewidth,trim={3cm 3cm 3cm 4cm},clip]{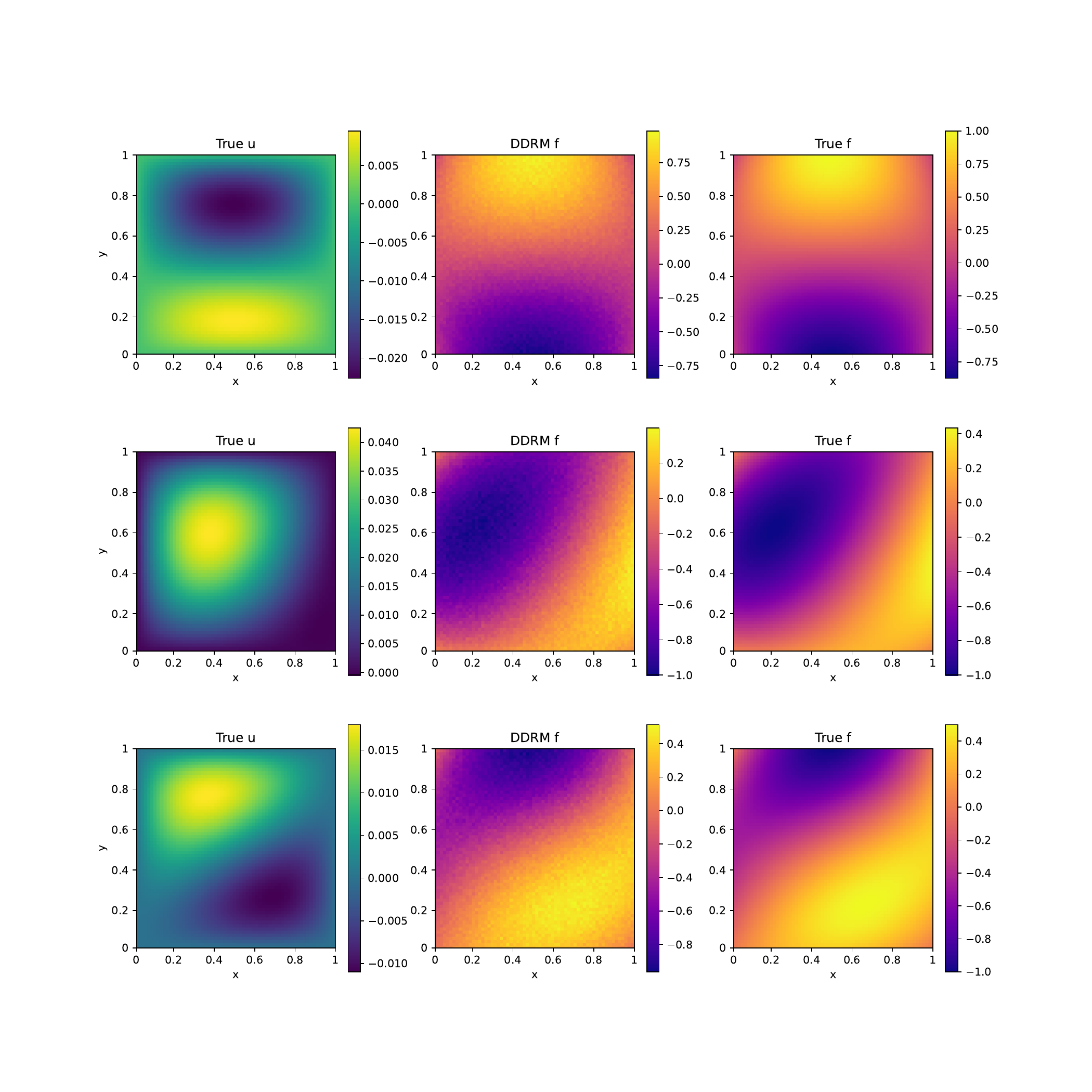}
        \caption{Plots of DDRM generated solutions of the inverse process $f(x,y)$ (middle), $u(x,y)$ (left), and the true $f(x,y)$ (right).}\label{fig:ddrm_f}
\end{figure}

In this section, we estimate the parameter $f(x,y)$ while keeping the $u(x,y)$ channel fixed using DDRM. We tested this on different randomly generated neural network functions explained in Section \ref{sec:nn_data}, three of which are plotted in Figure \ref{fig:ddrm_f}.

The plots show that the DDRM produces a significantly better approximation of the solution to the Poisson equation. The average MAE over 1024 different test samples is $3.215\times 10^{-2}$.

\section{Unconditionally Generated Data}\label{app:gen1}

\begin{figure}[H]
    \centering
    \includegraphics[width=\linewidth,trim={3cm 3cm 3cm 4cm},clip]{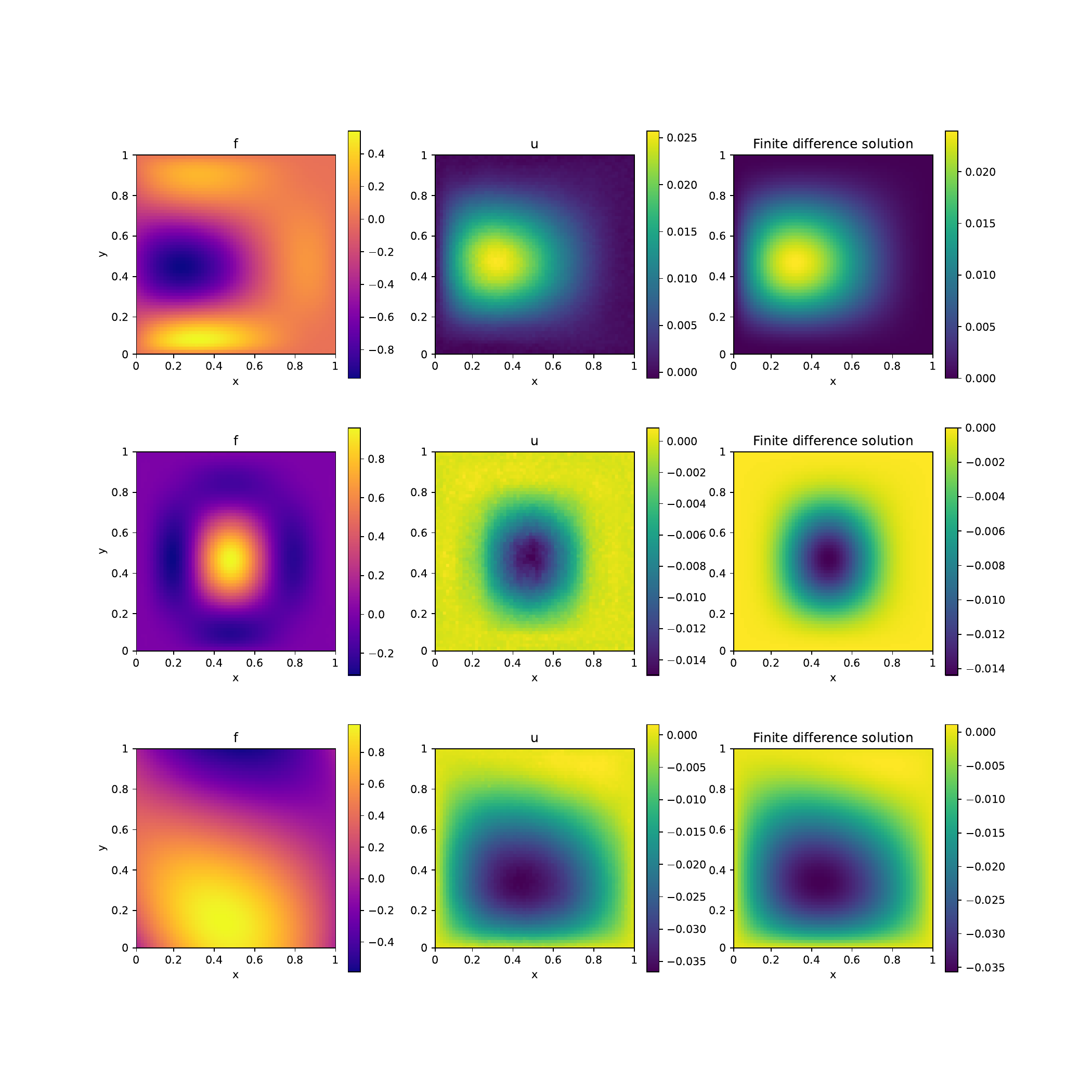}
        \caption{Plots of unconditionally generated pairs of $f(x,y)$ (left), $u(x,y)$ (middle), and the finite difference solution of $u(x,y)$ (right).}\label{fig:gen_figs1}
\end{figure}

We plotted three examples of solutions generated by our model unconditionally in Figure \ref{fig:gen_figs1}. The MAE between the generated solution $u(x,y)$ and the finite difference solution is $4.373\times 10^{-4}$, thus showing that the generated solutions are a good approximation to the true solution.

\end{document}